\UseRawInputEncoding
\documentclass{article}

\usepackage{microtype}
\usepackage{graphicx}
\usepackage{subfigure}
\usepackage{booktabs} 
\usepackage{makecell}
\usepackage{listings}
\usepackage{xcolor}  
\usepackage{mdframed}

\usepackage{listings}

\usepackage{hyperref}

\usepackage[preprint]{icml2025}

\usepackage{amsmath}
\usepackage{amssymb}
\usepackage{mathtools}
\usepackage{amsthm}

\usepackage[capitalize,noabbrev]{cleveref}

\theoremstyle{plain}

\theoremstyle{definition}

\theoremstyle{remark}

\usepackage[textsize=tiny]{todonotes}

\usepackage{multirow}
\usepackage{multicol}
\usepackage{enumitem}
\usepackage[most,skins,theorems]{tcolorbox}
\tcbset{
  aibox/.style={
    width=\linewidth,
    top=8pt,
    bottom=4pt,
    colback=blue!6!white,
    colframe=black,
    colbacktitle=black,
    enhanced,
    center,
    attach boxed title to top left={yshift=-0.1in,xshift=0.15in},
    boxed title style={boxrule=0pt,colframe=white,},
  }
}

\newtcolorbox{AIbox}[2][]{aibox,title=#2,#1}

\begin{document}

\twocolumn[
\icmltitle{Demystifying Long Chain-of-Thought Reasoning in LLMs}


\icmlsetsymbol{equal}{*}

\begin{icmlauthorlist}
\icmlauthor{Edward Yeo}{equal,inai}
\icmlauthor{Yuxuan Tong}{equal,thu}
\icmlauthor{Morry Niu}{inai}
\icmlauthor{Graham Neubig}{cmu}
\icmlauthor{Xiang Yue}{equal,cmu}
\end{icmlauthorlist}

\icmlaffiliation{cmu}{Carnegie Mellon University}
\icmlaffiliation{thu}{Tsinghua University. Work started when interning at CMU.}
\icmlaffiliation{inai}{IN.AI}

\icmlcorrespondingauthor{Xiang Yue}{xyue2@andrew.cmu.edu}

\icmlkeywords{Machine Learning, ICML}

\vskip 0.3in
]



\printAffiliationsAndNotice{\icmlEqualContribution} 

\begin{abstract}
Scaling inference compute enhances reasoning in large language models (LLMs), with long chains-of-thought (CoTs) enabling strategies like backtracking and error correction. Reinforcement learning (RL) has emerged as a crucial method for developing these capabilities, yet the conditions under which long CoTs emerge remain unclear, and RL training requires careful design choices. In this study, we systematically investigate the mechanics of long CoT reasoning, identifying the key factors that enable models to generate long CoT trajectories. Through extensive supervised fine-tuning (SFT) and RL experiments, we present four main findings: (1) While SFT is not strictly necessary, it simplifies training and improves efficiency; (2) Reasoning capabilities tend to emerge with increased training compute, but their development is not guaranteed, making reward shaping crucial for stabilizing CoT length growth; (3) Scaling verifiable reward signals is critical for RL. We find that leveraging noisy, web-extracted solutions with filtering mechanisms shows strong potential, particularly for out-of-distribution (OOD) tasks such as STEM reasoning; and (4) Core abilities like error correction are inherently present in base models, but incentivizing these skills effectively for complex tasks via RL demands significant compute, and measuring their emergence requires a nuanced approach. These insights provide practical guidance for optimizing training strategies to enhance long CoT reasoning in LLMs. Our code is available at: \href{https://github.com/eddycmu/demystify-long-cot}{https://github.com/eddycmu/demystify-long-cot}.
\end{abstract}

\section{Introduction}

Large language models (LLMs) \citep{brown2020language,
touvron2023llama,anthropic2023claude,openai2023gpt4} have demonstrated remarkable reasoning abilities in domains like mathematics \citep{cobbe2021gsm8k} and programming~\citep{chen2021humaneval}. A key technique for enabling reasoning abilities in LLMs is chain-of-thought (CoT) prompting \citep{wei2022cot}, which guides models to generate intermediate reasoning steps before arriving at a final answer.

Despite these advancements, LLMs still struggle with highly complex reasoning tasks, such as mathematical competitions \citep{hendrycks2021math}, PhD-level scientific QA \citep{rein2024gpqa}, and software engineering \citep{jimenez2024swebench}, even with CoT. Recently, OpenAI's o1 models \citep{openai2024o1} have demonstrated significant breakthroughs in these tasks. A key distinguishing feature of these models is their ability to scale up inference compute with long CoTs, which include strategies such as recognizing and correcting mistakes, breaking down difficult steps, and iterating on alternative approaches, leading to substantially longer and more structured reasoning processes.

Several efforts have attempted to replicate the performance of o1 models by training LLMs to generate long CoTs~\citep{qwen2024qwq, deepseekai2025r1, kimi2025k15, tinyzero, zeng2025simplerl}. Most of these approaches rely on verifiable rewards, such as accuracy based on ground-truth answers, which helps to avoid reward hacking in reinforcement learning (RL) at scale. However, a comprehensive understanding of how models learn and generate long CoTs remains limited. In this work, we systematically investigate the underlying mechanics of long CoT generation. Specifically, we explore:

1) \textit{Supervised fine-tuning (SFT) for long CoTs} -- the most direct way to enable long CoT reasoning. We analyze its scaling behavior and impact on RL, finding that long CoT SFT allows models to reach higher performance and also facilitates easier RL improvements than short CoT.

2) \textit{Challenges in RL-driven CoT scaling} -- we observe that RL does not always stably extend CoT length and complexity. To address this, we introduce a cosine length-scaling reward with a repetition penalty, which stabilizes CoT growth while encouraging emergent reasoning behaviors such as branching and backtracking.

3) \textit{Scaling up verifiable signals for long CoT RL} -- Verifiable reward signals are essential for stabilizing long CoT RL. However, scaling them up remains challenging due to the limited availability of high-quality, verifiable data. To address this, we explore the use of data containing noisy, web-extracted solutions~\cite{yue2024mammoth2}. While these ``silver'' supervision signals introduce uncertainty, we find that, with an appropriate mixture in SFT and filtration in RL, they show promise, especially in out-of-distribution (OOD) reasoning scenarios such as STEM problem-solving.

4) \textit{Origins of Long CoT Abilities and RL Challenges} – Core skills like branching and error validation are inherently present in base models, but effective RL-driven incentivization demands careful designs. We examine RL incentives on long CoT generation, trace reasoning patterns in pre-training data, and discuss nuances in measuring their emergence.

\section{Problem Formulation}
\label{sec:problemform}

In this section, we define the notation, followed by an overview of SFT and RL methods for eliciting long CoTs.

\begin{tcolorbox}[colback=lightgray!10, colframe=black, title={Research Aim}]
Our goal is to \textit{demystify long chain-of-thought reasoning} in LLMs. Through systematic analysis and ablations, we extract key insights and offer practical strategies to enhance and stabilize its performance.
\end{tcolorbox}
\vspace{-10pt}

\subsection{Notation}
\label{sec:notation}

Let \(x\) be a query, and let \(y\) be the output sequence. We consider a LLM parameterized by \(\theta\), which defines a conditional distribution over output tokens: $
\pi_\theta(y_t \mid x, y_{1:t-1}).
$

We denote by \(\text{CoT}(y)\subseteq y\) the tokens in the generated output that constitute the \emph{chain-of-thought}, which is often a reasoning trace or explanatory sequence. The final “answer” can be a separate set of tokens or simply the last part of \(y\).

In this work, we use the term \emph{long chain-of-thought} (\emph{long CoT}) to describe an extended sequence of reasoning tokens that not only exhibits a larger-than-usual token length but also demonstrates more sophisticated behaviors such as:

\noindent\textbf{1) Branching and Backtracking}: The model systematically explores multiple paths (branching) and reverts to earlier points if a particular path proves wrong (backtracking).

\noindent\textbf{2) Error Validation and Correction}: The model detects inconsistencies or mistakes in its intermediate steps and takes corrective actions to restore coherence and accuracy.

\subsection{Supervised Fine-Tuning (SFT)}
\label{sec:sft}

A common practice is to initialize the policy \(\pi_\theta\) via SFT~\citep{lamb2016professorforcing} on a dataset \(\mathcal{D}_{\text{SFT}} = \{(x_i, y_i)\}_{i=1}^N\), where \(y_i\) can be normal or long CoT reasoning tokens.

\subsection{Reinforcement Learning (RL)}
\label{sec:rl-formulation}

After optional SFT initialization, we can further optimize the generation of long CoT with reinforcement learning.

\noindent\textbf{Reward Function.} 
We define a scalar reward \(r_t\) designed to encourage correct and verifiable reasoning. We only consider the outcome-based reward for the final answer produced, and do not consider process-based reward for the intermediate steps. We denote
the term \(r_{\text{answer}}(y)\) to capture the correctness of the final solution.

\noindent\textbf{Policy Update.} We adopted Proximal Policy Optimization (PPO) \citep{schulman2017ppo} as the default policy optimization method in our experiments. We also briefly discuss
REINFORCE \citep{sutton2018reinforce} method in \autoref{result:reward-reinforce}. We adopt a rule-based verifier as the reward function, which compares the predicted answer with the ground truth answer directly. The resulting updates push the policy to generate tokens that yield higher reward.

\subsection{Training Setup}

We adopt \texttt{Llama-3.1-8B}~\cite{meta2023llama3} and \texttt{Qwen2.5} \texttt{-7B-Math}~\cite{qwen2024qwen25math} as the base models, which are representative general and math-specialized models respectively. For both SFT and RL, we use the 7,500 training sample prompt set of MATH \citep{hendrycks2021math} by default, with which verifiable ground truth answers are provided. For SFT when ground truth answers are available, we synthesize responses by rejection sampling \citep{zelikman2022star, dong2023raft, yuan2023rft, gulcehre2023rest, singh2023restem,yue2024mammoth, tong2024dartmath}. Specifically, we first sample a fixed number $N$ of candidate responses per prompt and then filter by only retaining ones with final answers consistent with the corresponding ground truth answers. We also discuss data like WebInstruct \cite{yue2024mammoth2} that is more diverse but without gold supervision signals like ground truth answers in \textsection\ref{sec:silver-data}. We train the models with the OpenRLHF framework~\cite{hu2024openrlhfeasytousescalablehighperformance}.

\subsection{Evaluation Setup}\label{sec:eval-setup}

We focus on four representative reasoning benchmarks: MATH-500, AIME 2024, TheoremQA \citep{chen2023theoremqa}, and MMLU-Pro-1k \citep{wang2024mmlupro}. Given that our training data is primarily in the mathematical domain, these benchmarks provide a comprehensive framework for both in-domain (MATH-500 test set) and out-of-domain evaluations (AIME 2024, TheoremQA, MMLU-Pro-1k). By default, we generate from the models using a temperature of $t = 0.7$, a top-$p$ value of 0.95, and a maximum output length of 16,384 tokens. Please refer to Appendix \ref{app:eval-setup} for further details on the evaluation setup.

\begin{figure*}[t!]
    \centering
    \includegraphics[width=0.96\linewidth]{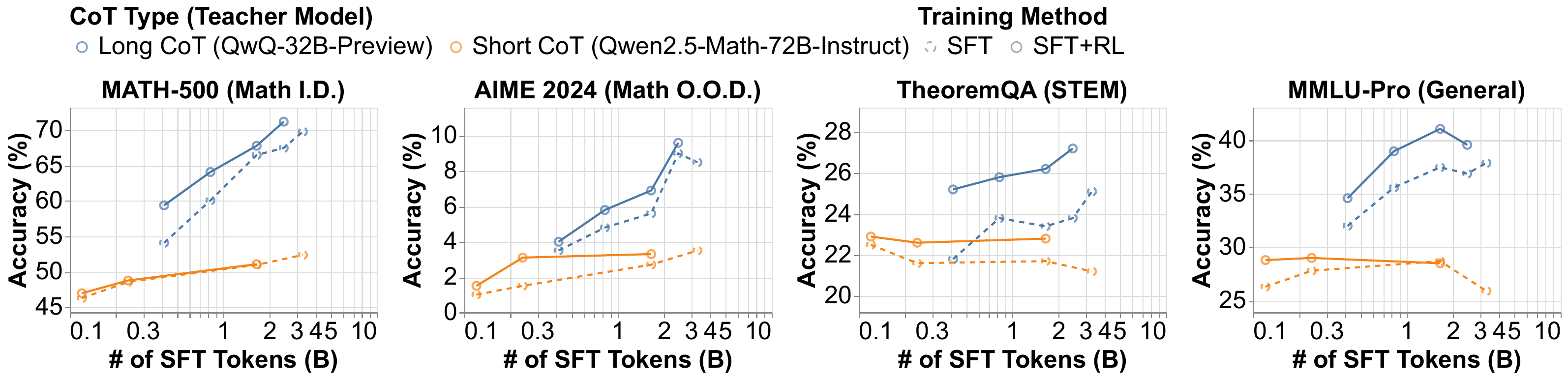}
    \vspace{-6pt}
    \caption{Scaling curves of SFT and RL on \texttt{Llama-3.1-8B}  with long CoTs and short CoTs. SFT with long CoTs can scale up to a higher upper limit and has more potential to further improve with RL.}
        \vspace{-10pt}

    \label{fig:cot-scaling}
\end{figure*}

\section{Impact of SFT on Long CoT}

In this section, we compare long and short CoT data for SFT and in the context of RL initialization.

\subsection{SFT Scaling}\label{subsec:sft-scaling}

To compare long CoT with short CoT, the first step is to equip the model with the corresponding behavior. The most straightforward approach is to fine-tune the base model on CoT data. Since short CoT is common, curating SFT data for it is relatively simple via rejection sampling from existing models. However, how to obtain high-quality long CoT data remains an open question. 

\noindent\textbf{Setup.} To curate the SFT data, for long CoT, we distill from \texttt{QwQ-32B-Preview} (we discuss other long CoT data construction methods in \textsection\ref{sec:long-cot-pattern}). For short CoT, we distill from \texttt{Qwen2.5-Math-72B-Instruct}, which is a capable short CoT model in math reasoning. Specifically, we perform rejection sampling by first sampling $N$ candidate responses per prompt and then filtering for ones with correct answers. For long CoT, we use $N \in \{32, 64, 128, 192, 256\}$, while for short CoT, we use $N \in \{32, 64, 128, 256\}$, skipping one $N$ for efficiency. 
In each case, the number of SFT tokens is proportional to $N$. We use the base model \texttt{Llama-3.1-8B} \citep{meta2023llama3}. Please refer to Appendix \ref{app:sft-setup} for more details about the SFT setup.

\noindent\textbf{Result.} The dashed lines in Figure \ref{fig:cot-scaling} illustrate that as we scale up the SFT tokens, long CoT SFT continues to improve model accuracy, whereas short CoT SFT saturates early at a lower accuracy level. For instance, on MATH-500, long CoT SFT achieves over 70\% accuracy and has yet to plateau even at 3.5B tokens. In contrast, short CoT SFT converges below 55\% accuracy, with an increase in SFT tokens from approximately 0.25B to 1.5B yielding only a marginal absolute improvement of about 3\%.

\begin{AIbox}{Takeaway 3.1 for SFT Scaling Upper Limit}
SFT with long CoT can scale up to a higher performance upper limit than short CoT. (Figure \ref{fig:cot-scaling})
\end{AIbox}

\begin{figure*}[!t]
    \centering
    \includegraphics[width=1\linewidth]{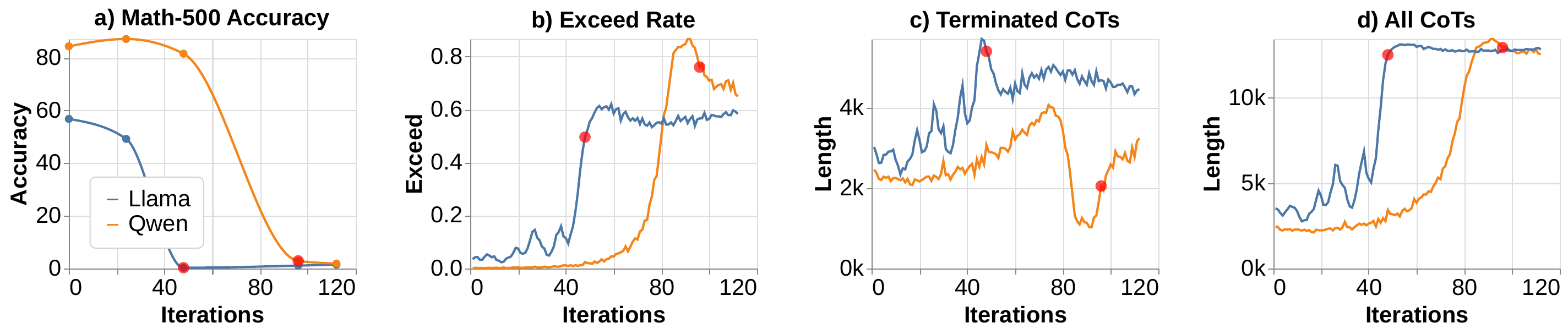}
    \vspace{-20pt}
    \caption{Both \texttt{Llama3.1-8B} and \texttt{Qwen2.5-Math-7B} models trained under RL with the Classic Reward manifested emergent CoT length scaling past the context window size, resulting in a decline in MATH-500 accuracy. The red points on the charts correspond to the iteration where the accuracy dropped to near zero. ``Terminated CoTs'' refer to responses that conclude within the context length.}
        \vspace{-10pt}

    \label{fig:reward-qwen-llama}
\end{figure*}

\subsection{SFT Initialization for RL}\label{sec:sft-init-for-rl}

Since RL is reported to have a higher upper limit than SFT, we compare long CoT and short CoT as different SFT initialization approaches for RL.

\noindent\textbf{Setup.}
We initialize RL using SFT checkpoints from \textsection\ref{subsec:sft-scaling}, and train for four epochs, sampling four responses per prompt. Our approach employs PPO \citep{schulman2017ppo} with a rule-based verifier from the MATH dataset, using its training split as our RL prompt set.
We adopt our cosine length scaling reward with the repetition penalty, which will be detailed in \textsection\ref{sec:reward-design}.
Further details about our RL setup and hyperparameters can be found in Appendix \ref{app:rl-setup} \& \ref{app:exp-hyperparams-sft-init-for-rl} respectively.

\noindent\textbf{Result.} The gap between solid and dashed lines in Figure \ref{fig:cot-scaling} shows that models initialized with long CoT SFT can usually be further significantly improved by RL, while models initialized with short CoT SFT see little gains from RL. For example, on MATH-500, RL can improve long CoT SFT models by over 3\% absolute, while short CoT SFT models have almost the same accuracies before and after RL.

\begin{AIbox}{Takeaway 3.2 for SFT Initialization for RL}
SFT with long CoTs makes further RL improvement easier, while short CoTs do not. (Figure \ref{fig:cot-scaling})
\end{AIbox}

\subsection{Sources of Long CoT SFT Data}\label{sec:long-cot-pattern}

To curate long CoT data, we compare two approaches: (1) \textbf{Construct} long CoT trajectories by prompting short CoT models to generate primitive actions and sequentially combining them; (2) \textbf{Distill} long CoT trajectories from existing long CoT models that exhibit emergent long CoT patterns.

\noindent\textbf{Setup.} To construct long CoT trajectories, we developed an Action Prompting framework (Appendix \ref{app:action-prompting}) which defined the following primitive actions: \texttt{clarify}, \texttt{decompose}, \texttt{solution\_step}, \texttt{reflection}, and \texttt{answer}. We employed multi-step prompting with a short CoT model (e.g., \texttt{Qwen2.5-72B-Instruct}) to sequence these actions, while a stronger model, \texttt{o1-mini-0912}, generates reflection steps incorporating self-correction. For distilling long CoT trajectories, we use \texttt{QwQ-32-Preview} as the teacher model. In both approaches, we adopt the MATH training set as the prompt set and apply rejection sampling. To ensure fairness, we use the same base model (\texttt{Llama-3.1-8B}), maintain approximately 200k SFT samples, and use the same RL setup as in \textsection\ref{sec:sft-init-for-rl}.

\noindent\textbf{Result.} Table \ref{tab:constructed-underperforms-emergent} shows that the model distilled from emergent long CoT patterns generalizes better than the constructed pattern, and can be further significantly improved with RL, while the model trained on constructed patterns cannot. Models trained with the emergent long CoT pattern achieve significantly higher accuracies on OOD benchmarks AIME 2024 and MMLU-Pro-1k, improving by 15-50\% relatively. 
Besides, on the OOD benchmark TheoremQA, RL on the long CoT SFT model significantly improves its accuracy by around 20\% relative, while the short CoT model's performance does not change. 
This is also why we conduct most of our experiments based on distilled long CoT trajectories.

\begin{AIbox}{Takeaway 3.3 for Long CoT Cold Start}
SFT initialization matters: high-quality, emergent long CoT patterns lead to significantly better generalization and RL gains. (Table \ref{tab:constructed-underperforms-emergent})
\end{AIbox}

\begin{table}[htbp]
\vspace{-15pt}
\caption{Emergent long CoT patterns outperform constructed ones. All the models here are fine-tuned from the base model \texttt{Llama-3.1-8B} with the MATH training prompt set.
}
\label{tab:constructed-underperforms-emergent}
\vskip 0.1in
\centering
\small
\begin{tabular}{@{}llcccc@{}}
\toprule
Training & Long CoT & MATH & AIME & Theo. & MMLU \\
Method & SFT Pattern & 500 & 2024 & QA & Pro-1k \\
\midrule
\multirow{2}{*}{SFT} & Constructed & 48.2 & 2.9 & 21.0 & 18.1 \\
& Emergent & \textbf{54.1} & \textbf{3.5} & \textbf{21.8} &\textbf{ 32.0} \\
\midrule
\multirow{2}{*}{SFT+RL} & Constructed & 52.4 & 2.7 & 21.0 & 19.2 \\
& Emergent & \textbf{59.4} & \textbf{4.0} & \textbf{25.2} & \textbf{34.6} \\
\bottomrule
\end{tabular}
\end{table}


\section{Impact of Reward Design on Long CoT}\label{sec:reward-design}

This section examines reward function design, with a focus on its influence on CoT length and model performance.

\subsection{CoT Length Stability}
\label{result:reward-length-stability}

Recent studies on long CoT \cite{deepseekai2025r1, kimi2025k15, hou2025advancinglanguagemodelreasoning} suggest that models naturally improve in reasoning tasks with increased thinking time. Our experiments confirm that models fine-tuned on long CoT distilled from \texttt{QwQ-32B-Preview} tend to extend CoT length under RL training, albeit sometimes unstably. This instability, also noted by \citet{kimi2025k15, hou2025advancinglanguagemodelreasoning}, has been addressed using techniques based on length and repetition penalties to stabilize training.

 \noindent\textbf{Setup.} We used two different models fine-tuned on long CoT data distilled from \texttt{QwQ-32B-Preview} using the MATH train split, with a context window size of 16K. The models were \texttt{Llama3.1-8B} and \texttt{Qwen2.5-Math-7B}. We used a rule-based verifier along and a simple reward of 1 for correct answers. We shall refer to this as the \textit{Classic Reward}. More details can be found in Appendix \ref{app:exp-hyperparams-reward-length-stability}.

\noindent\textbf{Results.} We observed that both models increased their CoT length during training, eventually reaching the context window limit. This led to a decline in training accuracy due to CoTs exceeding the allowable window size. Additionally, different base models exhibited distinct scaling behaviors. The weaker \texttt{Llama-3.1-8B} model showed greater fluctuations in CoT length compared to \texttt{Qwen-2.5-Math-7B}, as illustrated in Figure \ref{fig:reward-qwen-llama}.

We also found that the rate at which CoTs exceeded the context window size leveled off at a certain threshold below 1 (Figure \ref{fig:reward-qwen-llama}). This suggests that exceeding the limit started to apply significant downward pressure on the CoT length distribution, and highlights the context window size's role in implicit length penalization. Notably, a trajectory might be penalized even without an explicit exceed-length penalty due to reward or advantage normalization, both of which are standard in RL frameworks.

\begin{AIbox}{Takeaway \hypersetup{hidelinks}\ref{result:reward-length-stability} for CoT Length Stability}
CoT length does not always scale up in a stable fashion. (Figure \ref{fig:reward-qwen-llama}) 
\end{AIbox}

\begin{figure}[tbp]
    \centering
    \includegraphics[width=1\linewidth]{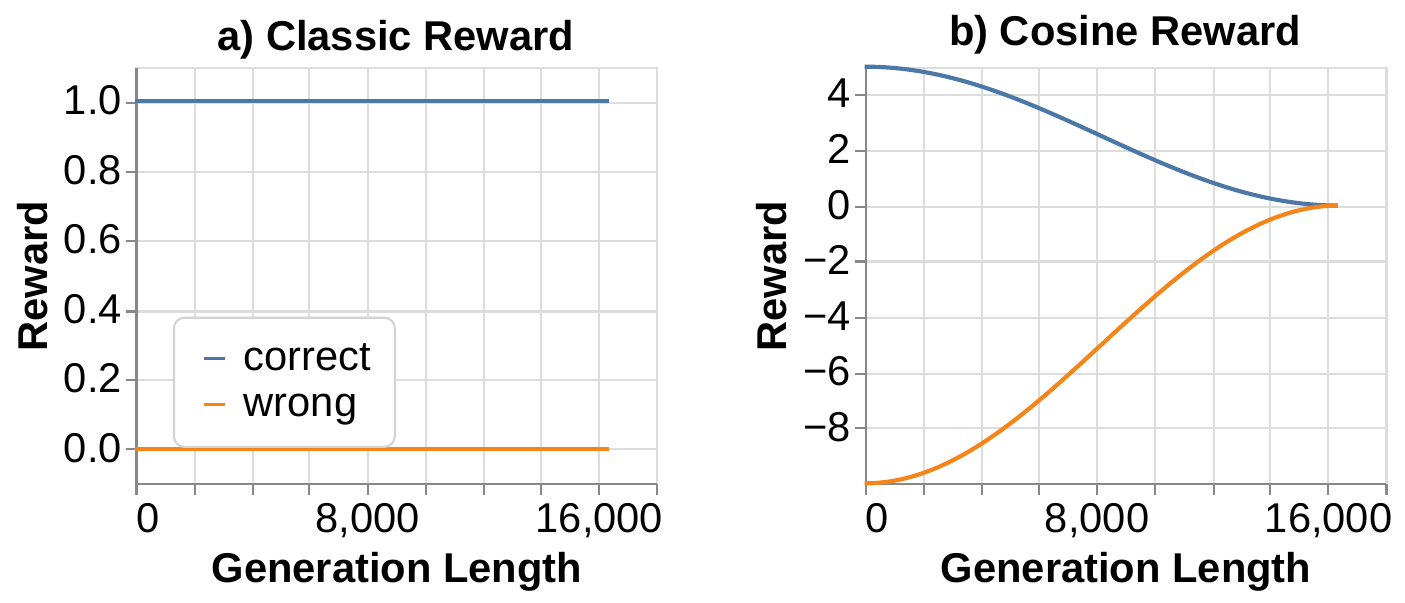}
    \vspace{-20pt}
    \caption{The Classic and Cosine Reward functions. The Cosine Reward varies with generation length.}
    \vspace{-10pt}
    \label{fig:reward-cosine}
\end{figure}

\subsection{Active Scaling of CoT Length}
\label{result:reward-length-scaling}

We found that reward shaping can be used to stabilize emergent length scaling. We designed a reward function to use CoT length as an additional input and to observe a few ordering constraints. Firstly, correct CoTs receive higher rewards than wrong CoTs. Secondly, shorter correct CoTs receive higher rewards than longer correct CoTs, which incentivizes the model to use inference compute efficiently. Thirdly, shorter wrong CoTs should receive higher penalties than longer wrong CoTs. This encourages the model to extend its thinking time if it is less likely to get the correct answer.

We found it convenient to use a piecewise cosine function, which is easy to tune and smooth. We refer to this reward function as the \textit{Cosine Reward}, visualized in Figure \ref{fig:reward-cosine}. This is a \textit{sparse} reward, only awarded once at the end of the CoT based on the correctness of the answer. The formula of \textbf{CosFn} can be found in equation \ref{eqn:cosine-lr} in the appendix.

\vspace{-10pt}
\begin{equation*}
\small
\label{eqn:reward-cosine}
\begin{aligned}
& R(C, L_{\text{gen}}) = 
&\begin{cases} 
    \text{CosFn}(L_{\text{gen}}, L_{\text{max}}, r_0^c, r_L^c),  & \text{if } C = 1, \\
    \text{CosFn} (L_{\text{gen}}, L_{\text{max}}, r_0^w, r_L^w),  & \text{if } C = 0, \\
    r_e, &\text{if } L_{\text{gen}} = L_{\text{max}}.
\end{cases}
\end{aligned}
\end{equation*}
\vspace{-25pt}

{\small
\begin{flalign*}
&\textbf{Hyperparameters:} && \\
&\quad r_0^c / r_0^w: \text{Reward (correct/wrong) for } L_{\text{gen}} = 0, && \\
&\quad r_L^c/r_L^w: \text{Reward (correct/wrong) for } L_{\text{gen}} = L_{\text{max}}, \\
&\quad r_e: \text{Exceed length penalty}, \\
&\textbf{Inputs:} && \\
&\quad C: \text{Correctness (0 or 1)}, && \\
&\quad L_{\text{gen}}: \text{Generation length.} &&
\end{flalign*}
}

\noindent\textbf{Setup.} We ran experiments with the Classic Reward and the Cosine Reward. We used the \texttt{Llama3.1-8B} fine-tuned on long CoT data distilled from \texttt{QwQ-32B-Preview} using the MATH train split, as our starting point. For more details, see Appendix \ref{app:exp-hyperparams-reward-length-scaling}.

\begin{figure}[tbp]
    \centering
    \includegraphics[width=1\linewidth]{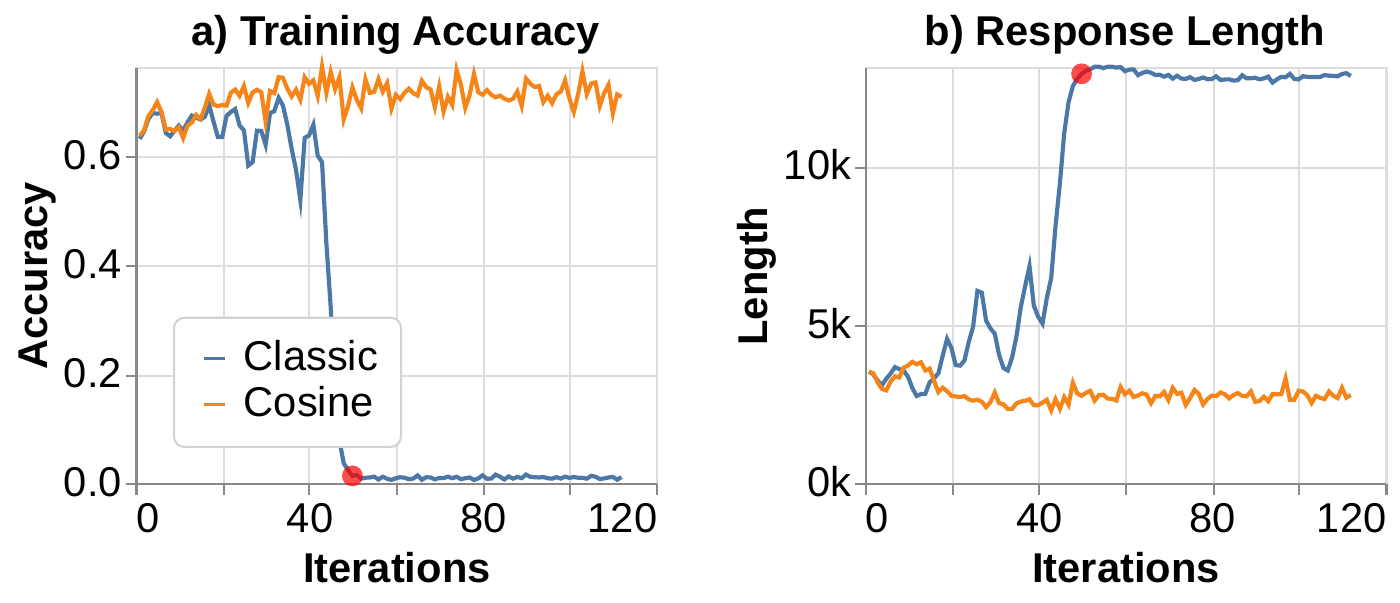}
    \vspace{-20pt}
    \caption{\texttt{Llama3.1-8B} trained with length shaping using the Cosine Reward exhibited more stable (a) training accuracy and (b) response length. This stability led to improved performance on downstream tasks (Figure \ref{fig:reward-eval}). Red points on the charts indicate iterations where training accuracy dropped to near zero.}
    \vspace{-10pt}
    \label{fig:reward-llama-classic}
\end{figure}

\begin{figure*}[tbp]
    \centering
    \includegraphics[width=1\linewidth]{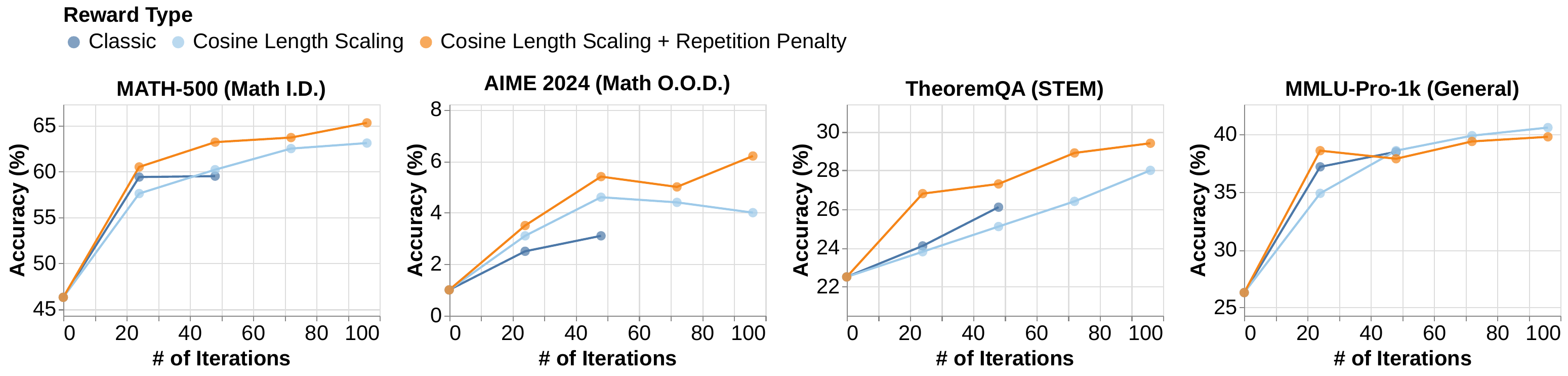}
    \vspace{-20pt}
    \caption{Performance of \texttt{Llama-3.1-8B} trained with different reward functions on a variety of evaluation benchmarks.}
    \vspace{-10pt}
    \label{fig:reward-eval}
\end{figure*}

\noindent\textbf{Result.} We found that the Cosine Reward significantly stabilized the length scaling behavior of the models under RL, thereby also stabilizing the training accuracy and improving RL efficiency (Figure \ref{fig:reward-llama-classic}). We also observed improvements in model performance on downstream tasks (Figure \ref{fig:reward-eval}).

\begin{AIbox}{Takeaway \hypersetup{hidelinks}\ref{result:reward-length-scaling} for Active Scaling of CoT Length}
Reward shaping can be used to stabilize and control CoT length while improving accuracy. (Figure \ref{fig:reward-llama-classic}, \ref{fig:reward-eval})
\end{AIbox}

\subsection{Cosine Reward Hyperparameters}
\label{result:reward-cosine-hyperparams}

The Cosine Reward hyperparameters can be tuned to shape CoT length in different ways.

\noindent\textbf{Setup.} We set up RL experiments with the same model fine-tuned on long CoT distilled from \texttt{QwQ-32B-Preview}, but with different hyperparameters for the Cosine Reward function. We tweaked the correct and wrong rewards $r_0^c, r_L^c, r_0^w, r_L^w$ and observed their impact on the CoT lengths. For more details, see Appendix \ref{app:exp-hyperparams-reward-cosine-hyperparams}.

\noindent\textbf{Result.} We see from Figure \ref{fig:reward-cosine-hyperparams} in the Appendix that if the reward for a correct answer increases with CoT length ($r_0^c < r_L^c$), the CoT length increases explosively. We also see that the lower the correct reward relative to the wrong reward, the longer the CoT length. We interpret this as a kind of trained risk aversion, where the ratio of the correct and wrong rewards determines how confident the model has to be about an answer for it to derive a positive expected value from terminating its CoT with an answer.

\begin{AIbox}{Takeaway \hypersetup{hidelinks}\ref{result:reward-cosine-hyperparams} for Cosine Reward Hyperparameters}
Cosine Reward can be tuned to incentivize various length scaling behaviors. (Appendix Figure \ref{fig:reward-cosine-hyperparams})
\end{AIbox}

\subsection{Context Window Size}
\label{result:reward-context-window}

We know that longer contexts give a model more room to explore, and with more training samples, the model eventually learns to utilize more of the context window. This raises an interesting question -- are more training samples necessary to learn to utilize a larger context window?

\noindent\textbf{Setup.} We set up 3 experiments using the same starting model fine-tuned on long CoT data distilled from \texttt{QwQ-32B-Preview} with the MATH train split. We also used the latter as our RL prompt set. Each ablation used the Cosine Reward and repetition penalty with a different context window size (4K, 8K, and 16K). For more details, see Appendix \ref{app:exp-hyperparams-reward-context-window}.

\begin{figure*}
    \centering
    \includegraphics[width=1\linewidth]{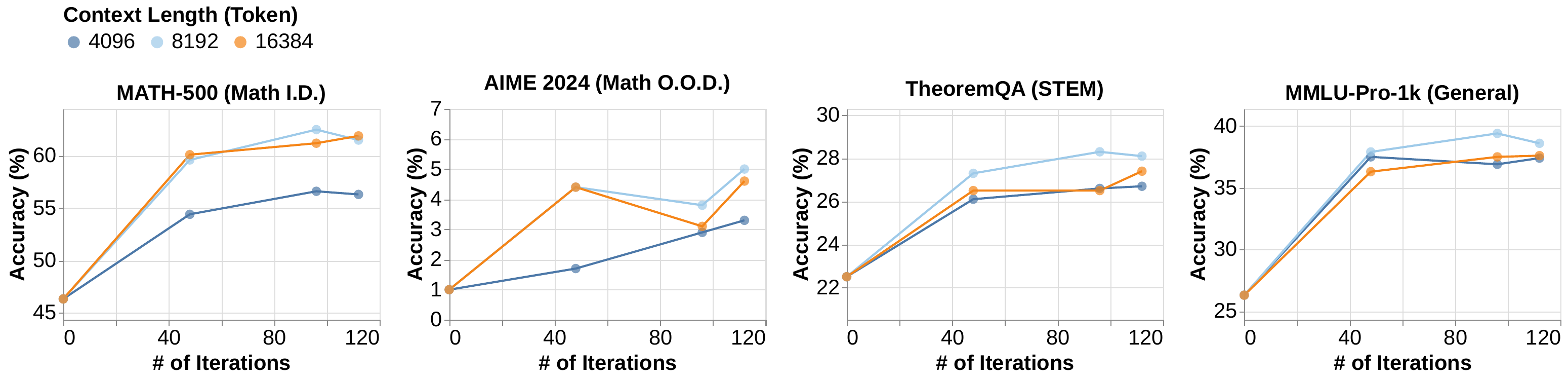}
    \vspace{-20pt}
    \caption{Performance of \texttt{Llama-3.1-8B} trained with different context window sizes. All experiments used the same number of training samples.}
    \vspace{-10pt}
    \label{fig:reward-context}
\end{figure*}

\noindent\textbf{Result.} We found that the model with a context window size of 8K performed better than the model with 4K, as expected. However, we observed performance was better under 8K than 16K. Note that all three experiments used the same number of training samples (Figure \ref{fig:reward-context}). We see this as an indication that models need more training compute to learn to fully utilize longer context window sizes, which is consistent with the findings of \cite{hou2025advancinglanguagemodelreasoning}.

\begin{AIbox}{Takeaway \hypersetup{hidelinks}\ref{result:reward-context-window} for Context Window Size}
Models might need more training samples to learn to utilize larger context window sizes. (Figure \ref{fig:reward-context})
\end{AIbox}

\subsection{Length Reward Hacking}
\label{result:reward-hacking}

We observed that with enough training compute, the model started to show signs of reward hacking, where it increased the lengths of its CoTs on hard questions using repetition rather than learning to solve them. We also noted a fall in the branching frequency of the model, which we estimated by counting the number of times the pivot keyword "\texttt{alternatively,}" appeared in the CoT (Figure \ref{fig:reward-cosine-branching}).

We mitigated this by implementing a simple $N$-gram repetition penalty (Algorithm \ref{alg:reward-repetition-penalty}). We observed that the penalty was most effectively applied on repeated tokens, rather than as a sparse reward for the entire trajectory. Similarly, we found that discounting the repetition penalty when calculating the return was effective. Specific feedback about where the repetition occurred presumably made it easier for the model to learn not to do it (see more in \textsection\ref{result:optimal-discount}).

\noindent\textbf{Setup.} We used the \texttt{Llama3.1-8B} model fine-tuned on long CoT data distilled from \texttt{QwQ-32B-Preview}. We ran two RL training runs, both using the Cosine Reward, but with and without the repetition penalty. For more details, please refer to Appendix \ref{app:exp-hyperparams-reward-hacking}.

\noindent\textbf{Result.} The repetition penalty resulted in better downstream task performance and also shorter CoTs, meaning there was better utilization of inference compute (Figure \ref{fig:reward-eval}).

\textbf{Observation.} Our experiments revealed a relationship between the repetition penalty, training accuracy, and the Cosine Reward. When training accuracy was low, the Cosine Reward exerted greater upward pressure on CoT length, leading to increased reward hacking through repetition. This, in turn, required a stronger repetition penalty. Future work could further investigate these interactions and explore dynamic tuning methods for better optimization.

\begin{AIbox}{Takeaway \hypersetup{hidelinks}\ref{result:reward-hacking} for Length Reward Hacking}
Length rewards will be hacked with enough compute (Figure \ref{fig:reward-cosine-branching}), but this can be mitigated using a repetition penalty. (Figure \ref{fig:reward-eval})
\end{AIbox}

\subsection{Optimal Discount Factors}
\label{result:optimal-discount}

We hypothesized that applying the repetition penalty with temporal locality (i.e., a low discount factor) would be most effective, as it provides a stronger learning signal about the specific offending tokens. However, we also observed performance degradation when the discount factor for the correctness (cosine) reward was too low.

To optimally tune both reward types, we modified the GAE formula in PPO to accommodate multiple reward types, each with its own discount factor $\gamma$: $\hat{A}_t = \sum_{l=0}^{L} \sum_{m}^{M}\gamma_{m}^l r_{m, t + l} - V(s_t)$. For simplicity, we set $\lambda = 1$, which proved effective, though we did not extensively tune this parameter.

\noindent\textbf{Setup.} We ran multiple RL experiments with the same \texttt{Llama3.1-8B} model fine-tuned on \texttt{QwQ-32B-Preview} distilled long CoT data. We used the Cosine Reward and repetition penalty but with different combinations of discount factors. For more details, please see Appendix \ref{app:exp-hyperparams-optimal-discount}.

\noindent\textbf{Result.} A lower discount factor effectively enforces the repetition penalty, whereas a higher discount factor enhances the correctness reward and the exceed-length penalty. The higher factor allows the model to be adequately rewarded for selecting a correct answer earlier in the CoT (Figure \ref{fig:multiple-gamma}).

We observed a rather interesting phenomenon where decreasing the discount factor $\gamma$ of the correctness (cosine) reward increased the branching frequency in the model's CoT, making the model quickly give up on approaches that did not seem to lead to a correct answer immediately (Figure \ref{fig:reward-indecisive}, Extract in Appendix \ref{extract:reward-short-term}). We hypothesize that this short-term thinking was due to a relatively small number of tokens preceding the correct answer receiving rewards, which means stepping stones to the right answer are undervalued. Such behavior degraded performance (Figure \ref{fig:multiple-gamma}). However, we think this qualitative result might be of potential interest to the research community, due to its similarity to the relationship between behaviors like delayed gratification and the distribution of rewards given to the biological brain \cite{doi:10.1126/sciadv.abg6611}.

\begin{AIbox}{Takeaway \hypersetup{hidelinks}\ref{result:optimal-discount} for Optimal Discount Factors}
Different kinds of rewards and penalties have different optimal discount factors. (Figure \ref{fig:multiple-gamma})
\end{AIbox}

\section{Scaling up Verifiable Reward}\label{sec:silver-data}

Verifiable reward signals like ones based on ground-truth answers are essential for stabilizing long CoT RL for reasoning tasks. However, it is difficult to scale up such data due to the limited availability of high-quality human-annotated verifiable data for reasoning tasks. As an attempt to counter this, we explore using other data that is more available despite more noise, like reasoning-related QA pairs extracted from web corpora. Specifically, we experiment with the WebInstruct dataset \citep{yue2024mammoth2}. For efficiency, we construct WebInstruct-462k, a deduplicated subset derived via MinHash \citep{broder1998minhash}. 

\subsection{SFT with Noisy Verifiable Data}\label{sec:sft-with-noisy-verifiable-data}

We first explore adding such diverse data to SFT. Intuitively, despite less reliable supervision signals, diverse data might facilitate the model’s exploration during RL.

\noindent\textbf{Setup.} We experiment with three setups, varying the proportion of data without gold supervision signals: 0\%, 100\%, and approximately 50\%. We conduct long CoT SFT by distilling from \texttt{QwQ-32B-Preview}. For data with gold supervision signals (MATH), ground truth answers are used for rejection sampling. In contrast, for data from WebInstruct without fully reliable supervision signals but with a much larger scale, we sample one response per prompt from the teacher model without filtration. For RL here, we adopt the same setup as in \textsection\ref{sec:sft-init-for-rl}, using the MATH training set.

\noindent\textbf{Result.} Table \ref{tab:diverse-silver-improve-general-reasoning} shows that incorporating silver-supervised data improves average performance. Adding WebInstruct data to long CoT SFT yields a substantial 5–10\% absolute accuracy gain on MMLU-Pro-1k over using MATH alone. Furthermore, mixing MATH and WebInstruct data achieves the best average accuracy across benchmarks.

\begin{table}[htbp]
\vspace{-10pt}
\caption{
Adding data with a silver supervision signal is often beneficial.
``WebIT'' is the abbreviation of WebInstruct.
}
\vspace{5pt}
\label{tab:diverse-silver-improve-general-reasoning}
\centering
\small
\resizebox{\linewidth}{!}{
\begin{tabular}{@{}llccccc@{}}
\toprule
Long CoT & Training & MATH & AIME & Theo. & MMLU & \multirow{2}{*}{AVG} \\
SFT Data & Method & 500 & 2024 & QA & Pro-1k \\
\midrule
\multirow{2}{*}{100\% MATH}
 & SFT & 54.1 & 3.5 & 21.8 & 32.0 & 27.9 \\
 & SFT + RL & \textbf{59.4} & 4.0 & \textbf{25.2} & 34.6 & 30.8 \\
\midrule
\multirow{2}{*}{100\% WebIT}
& SFT & 41.2 & 0.8 & 21.9 & 41.1 & 26.3 \\
& SFT + RL & 44.6 & 1.9 & 22.5 & \textbf{43.3} & 28.1 \\
\midrule
50\% MATH & SFT & 53.6 & \textbf{4.4} & 23.5 & 41.7 & 30.8 \\
+ 50\% WebIT & SFT + RL & 57.3 & 3.8 & 25.1 & 42.0 & \textbf{32.1} \\
\bottomrule
\end{tabular}
}
\end{table}

\begin{AIbox}{Takeaway 5.1 for SFT with Noisy Verifiable Data}
Adding noisy but diverse data to SFT leads balanced performance across different tasks. (Table \ref{tab:diverse-silver-improve-general-reasoning})
\end{AIbox}

\subsection{Scaling up RL with Noisy Verifiable Data}
\label{result:reward-verify-clean}

We compare two main approaches to obtain rewards from noisy verifiable data: 1) to extract short-form answers and use a rule-based verifier; 2) to use a model-based verifier capable of processing free-form responses. 

Here a key factor is whether the QA pair can have a short-form answer. So we also compare whether the dataset is filtered by only retaining samples with short-form answers. 

\noindent\textbf{Setup.}
We implement the model-based verifier by prompting \texttt{Qwen2.5-Math-7B-Instruct} with the raw reference solution.
To extract short-form answers, we first prompt \texttt{Llama-3.1-8B-Instruct} to extract from the raw responses and then apply rejection sampling with \texttt{QwQ-32B-Preview}. Specifically, we generate two responses per prompt from WebInstruct-462k and discard cases where neither response aligns with the extracted reference answers. This process yields approximately 189k responses across 115k unique prompts.
Our case studies show that the rejection sampling drops many prompts due to:
1) many WebInstruct prompts lack short-form answers that our rule-based verifier can process effectively,
and 2) some prompts are too difficult even for \texttt{QwQ-32B-Preview}.
For SFT we train \texttt{Llama-3.1-8B} on the filtered dataset as initialization for reinforcement learning (RL).
In the RL stage, we use the full 462k prompt set in the unfiltered setup and the 115k subset in the filtered setup, training with 30k prompts and 4 responses per prompt.
Further details about the model-based verifier, the answer extraction and the RL hyperparameters can be found in Appendix  \& \ref{app:exp-hyperparams-reward-verify-clean} \& \ref{app:model-based-verifier} \& \ref{app:ans-extract} respectively.

\begin{table}[htbp]
\vspace{-15pt}
\caption{Performance of RL with different verifiers and prompt filtering methods. All the models here are fine-tuned from \texttt{Llama-3.1-8B}. The ``MATH Baseline'' is the model trained with SFT and RL on MATH only in Table \ref{tab:diverse-silver-improve-general-reasoning}. The other models are trained with SFT by distillation from \texttt{QwQ-32B-Preview} and RL with different setups.}
\label{tab:verification-types}
\vskip 0.1in
\centering
\small
\resizebox{\linewidth}{!}{
\begin{tabular}{@{}llccccc@{}}
\toprule
Prompt & Verifier & MATH & AIME & Theo. & MMLU \\
Set & Type & 500 & 2024 & QA & Pro-1k \\
\midrule
\multicolumn{2}{c}{MATH Baseline} & 59.4 & 4.0 & 25.2 & 34.6 \\
\midrule
\multicolumn{2}{c}{SFT Initialization} & 46.6 & 1.0 & 23.0 & 28.3 \\
\midrule
\multirow{2}{*}{Unfiltered} & Rule-Based & 45.4 & 3.3 & 25.9 & 35.1 \\
 & Model-Based & 47.9 & 3.5 & 26.2 & 40.4 \\
\midrule
\multirow{2}{*}{Filtered} & Rule-Based & \textbf{48.6} & 3.3 & \textbf{28.1} & \textbf{41.4} \\
 & Model-Based & 47.9 & \textbf{3.8} & 26.9 & \textbf{41.4} \\
\bottomrule
\end{tabular}
}
\end{table}

\noindent\textbf{Result.} \autoref{tab:verification-types} shows that RL with the rule-based verifier on the filtered prompt set with short-form answers achieves the best performance across most benchmarks under the same number of RL samples. This might indicate that rule-based verifier after appropriate filtration can produce the highest-quality reward signals from noisy verifiable data.
Moreover, compared to the model trained on human-annotated verified data (MATH), leveraging noisy yet diverse verifiable data still significantly boosts performance on O.O.D. benchmarks, with absolute gains of up to 2.9\% on TheoremQA and 6.8\% on MMLU-Pro-1k. In contrast, applying a rule-based verifier to unfiltered data results in the worst performance.
This might be caused by its low training accuracy on free-form answers, while the model-based verifier achieves much better performance.

\begin{AIbox}{Takeaway \hypersetup{hidelinks}\ref{result:reward-verify-clean} for RL with Noisy Verifiable Data}
To obtain reward signals from noisy verifiable data, the ruled-based verifier after filtering the prompt set for short-form answers works the best. (Table \ref{tab:verification-types})
\end{AIbox}

\section{Exploration on RL from the Base Model}
\label{sec:rl-from-base}

DeepSeek-R1 \citep{deepseekai2025r1} has demonstrated that long chain-of-thought reasoning can emerge by scaling up reinforcement learning compute on a base model. Recent studies \citep{zeng2025simplerl, tinyzero} have attempted to replicate this progress by running a relatively small number of RL iterations to observe the emergence of long CoT behavior (e.g., the ``aha moment''~\citep{deepseekai2025r1}, an emergent realization moment that enables critical functions like self-validation and correction).
We also explore the method of RL from the base model in this section.

\subsection{Nuances in Analysis Based on Emergent Behaviors}
\label{result:base-reflections-existence}

Self-validation behaviors are sometimes flagged as emergent behaviors or ``aha-moment'' by the model's exploration, since such patterns are rare in short CoT data. However, we notice that sometimes self-validation behaviors already exist in the base model  and reinforcing them through RL requires strict conditions, such as a strong base model.

\noindent\textbf{Setup.}
We follow the setup from \citet{zeng2025simplerl} to train \texttt{Qwen2.5-Math-7B} using PPO  with a rule-based verifier on approximately 8k MATH level 3-5 questions, but we use our own rule-based verifier implementation. For inference, we adopt temperature $t = 0$ (greedy decoding), as our preliminary experiments show that $t=0$ usually significantly outperforms $t>0$ for models obtained by direct RL from \texttt{Qwen2.5-Math-7B}. We use the maximum output length of 4096 tokens considering the training context length of 4096 tokens. Note that we use zero-shot prompting for the base model to avoid introducing biases to the output pattern. We select five representative keywords, ``wait'', ``recheck'', ``alternatively'', ``retry'' and ``however'' from long CoT cases in previous works \citep{openai2024o1,deepseekai2025r1,tinyzero,zeng2025simplerl}, and calculate their frequencies to quantify the extent to which the model does self-validation. Further details about the RL hyperparameters can be found in Appendix \ref{app:exp-hyperparams-rl-from-base}.

\begin{figure*}[htbp]
    \centering
    \includegraphics[width=1\linewidth]{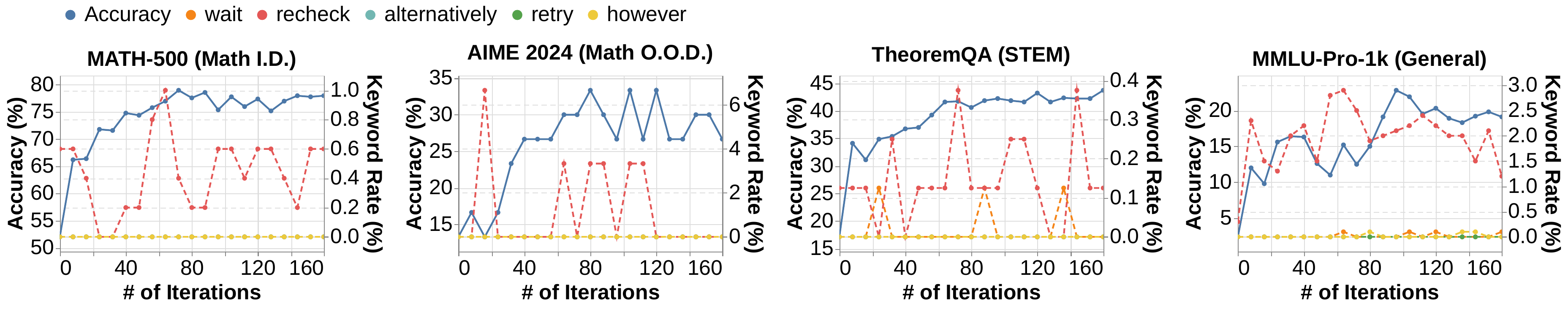}
    \vspace{-20pt}
    \caption{Dynamics of accuracies and reflection keyword rates on different benchmarks during our RL from the base model \texttt{Qwen2.5-Math-7B}. We do not see the keyword rates of ``wait'', ``alternatively'', and ``recheck'' get significantly improved during the RL training even though the accuracy is steadily increasing. }
    \label{fig:reflection-acc-keywords-rate}
\end{figure*}

\begin{figure*}[htbp]
    \centering
    \includegraphics[width=1\linewidth]{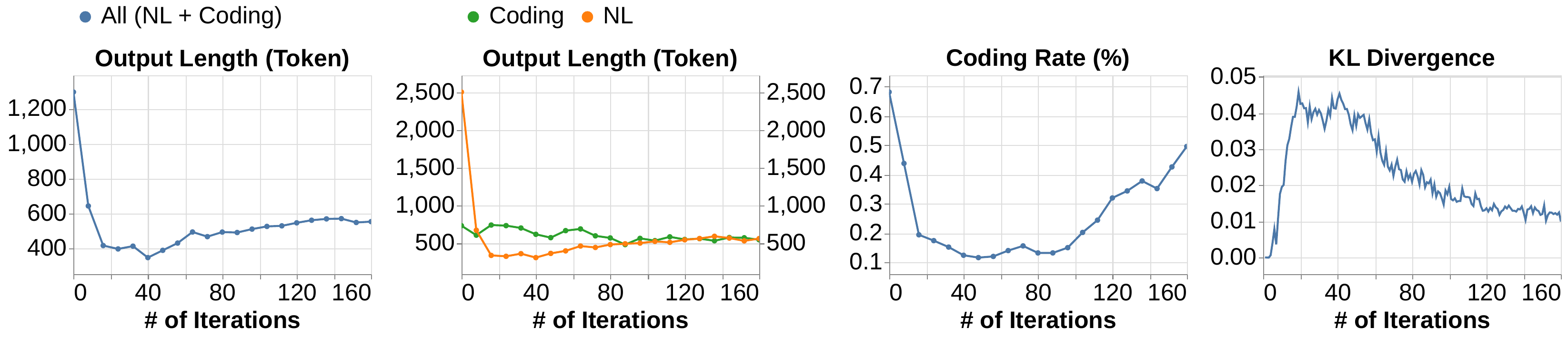}
    \vspace{-20pt}
    \caption{Dynamics of the output token lengths and the coding rate on MATH-500 and the KL divergence of the policy over the base model on MATH Lv3-5 (training data) during our RL from \texttt{Qwen2.5-Math-7B}.}
    \vspace{-10pt}
    \label{fig:code-rate-output-len}
\end{figure*}

\noindent\textbf{Result.} Figure \ref{fig:reflection-acc-keywords-rate} shows that our RL from \texttt{Qwen2.5-} \texttt{Math-7B} effectively boosts the accuracies, but does not increase the frequency of the ``recheck`` pattern existing in the output of the base model,
nor effectively incentivize other reflection patterns such as ``retry'' and ``alternatively''. This indicates that RL from the base model does not necessarily incentivize reflection patterns, though significantly boosting the performance. Sometimes such behaviors exist in the base model's output and RL does not substantially enhance them. So we might need to be more careful about recognizing emergent behaviors.


\subsection{Nuances in Analysis Based on Length Scaling}
\label{sec:analysis-length-scaling} 

The length scaling up is recognized as another important feature of the effective exploration of the model. However, we notice that sometimes length scaling up can be accompanied by the KL divergence decreasing, which raises the possibility that the length is influenced by the KL penalty and is just reverting back to the base model's longer outputs, rather than reflecting the acquisition of long CoT ability.

\noindent\textbf{Setup.} The setup is the same as in \textsection\ref{result:base-reflections-existence}. Besides the output token length, we also calculate the ``coding rate''. We classify the model's output as ``coding'' if it contains the ``\texttt{```python}'', since \texttt{Qwen2.5-Math-7B} uses both natural language and coding to solve mathematical problems. Note that the ``coding'' output here is actually a special form of natural language output, where the code in it is not executed, and the code's output is generated by the model.

\noindent\textbf{Result.} Figure \ref{fig:code-rate-output-len} (1) shows that the length of the output token increases after an initial drop, but never exceeds the initial length of the base model.

\citet{zeng2025simplerl} suggest that the initial drop may be due to the model transitioning from generating long coding outputs to shorter natural language outputs. However, Figure \ref{fig:code-rate-output-len} (2) indicates that natural language outputs are actually longer than coding outputs, and the initial drop in length occurs in both types of output. Furthermore, Figure \ref{fig:code-rate-output-len} (3) shows that the coding rate subsequently increases again, suggesting that the distinction between coding and natural language may not significantly impact the optimization process.

Moreover, we suspect that the subsequent length scaling up is not from the model's exploration, since when the length scales up, the KL divergence of the policy over the base model drops, as shown in Figure \ref{fig:code-rate-output-len} (4). This might indicate that it is the KL penalty influencing length. If that is the case, there is little potential for the policy output length to exceed the base model's since the exploration is limited by the KL constraint.


\subsection{Potential Reasons Why Emergent Behavior is Not Observed with \texttt{Qwen2.5-Math-7B}}

Our detailed analysis of RL from \texttt{Qwen2.5-Math-7B}, as presented in \textsection\ref{result:base-reflections-existence} and \textsection\ref{sec:analysis-length-scaling}, suggests that it fails to fully replicate the training behavior of \texttt{DeepSeek-R1}. We identify the following potential causes: 1) The base model, being relatively small (7B parameters), may lack the capacity to quickly develop such complex abilities when incentivized. 2) The model might have been overexposed to MATH-like short instruction data during (continual) pre-training and annealing, leading to overfitting and hindering the development of long CoT behaviors.

\subsection{Comparison between RL from the Base Model and RL from Long CoT SFT}
\label{sec:rl-from-base-vs-long-cot-sft}

We compare the performance of RL from the base model and RL from long CoT SFT and find that RL from long CoT SFT generally performs better.

\noindent\textbf{Setup.}
We compare using the base model \texttt{Qwen2.5-} \texttt{Math-7B}. The results of RL from the base model are from the model trained in \textsection\ref{result:base-reflections-existence}. For RL from long CoT SFT, we adopt a setup similar to \textsection\ref{sec:sft-init-for-rl}. Specifically, we choose the 7.5k MATH training set as the prompt set, curate the SFT data by rejection sampling with 32 candidate responses per prompt using \texttt{QwQ-32B-Preview}, and perform PPO using our cosine length-scaling reward with repetition penalty and our rule-based verifier, sampling 8 responses per prompt and training for 8 epochs. To adapt \texttt{Qwen2.5-Math-7B} with a pre-training context length of only 4096 tokens to long CoT SFT and RL, we multiply its RoPE \citep{su2024rope} $\theta$ by 10 times. We don't report the results of RL with classic reward from long CoT SFT since it collapses. For evaluation, we adopt our default temperature sampling setup for RL from long CoT SFT as in \textsection\ref{sec:eval-setup} and greedy decoding setup for RL from the base model as in \textsection\ref{result:base-reflections-existence} for the best performance. Further details about the distillation, SFT hyperparameters and RL hyperparameters can be found in Appendix \ref{app:distill} \& \ref{app:sft-setup} \& \ref{app:exp-hyperparams-rl-from-base}, respectively.

\begin{table}[htbp]
\caption{
Performance of different models based on \texttt{Qwen2.5-Math-7B}. The SFT data here is distilled with rejection sampling from \texttt{QwQ-32B-Preview}.
}
\label{tab:rl-from-base-vs-from-long-sft-rl}
\vskip 0.1in
\centering
\small
\begin{tabular}{@{}lccccc@{}}
\toprule
\multirow{2}{*}{Setup} & MATH & AIME & Theo. & MMLU & \multirow{2}{*}{AVG}\\
& 500 & 2024 & QA & Pro-1k \\
\midrule
Base (0-shot) & 52.0 & 13.3 & 17.1 & 2.4 & 21.2 \\
(Direct) RL & 77.4 & 23.3 & 43.5 & 19.7 & 41.0 \\
SFT  & 84.0 & 24.4 & 42.2 & 38.5 & 47.3 \\
SFT + RL & \textbf{85.9} & \textbf{26.9} & \textbf{45.4} & \textbf{40.6} & \textbf{49.7} \\
\bottomrule
\end{tabular}
\end{table}

\noindent\textbf{Result.} Table \ref{tab:rl-from-base-vs-from-long-sft-rl} shows that, on \texttt{Qwen2.5-Math-7B}, RL initialized from the long CoT SFT model significantly outperforms RL from the base model and further improves upon the long CoT SFT itself. Specifically, RL from long CoT SFT with our cosine reward surpasses RL from the base model by a substantial 8.7\% on average and improves over the SFT initialization by 2.6\%. Notably, simply applying SFT with long CoT distilled from \texttt{QwQ-32B-Preview} already yields strong performance.


\subsection{Long CoT Patterns in Pre-training Data}
\label{result:base-cot-origin}

Based on the results in \textsection\ref{result:base-reflections-existence}, we hypothesize that incentivized behaviors, such as the model revisiting its solutions, may have already been partially learned during pre-training. To examine this, we employed two methods to investigate whether such data are already present on the web.

Firstly, we used a generative search engine Perplexity.ai to identify webpages explicitly containing problem-solving steps that approach problems from multiple angles or perform verification after providing an answer. The query we used and the examples we identified are in Appendix \ref{webpage:explicit-revision-correct}).

Secondly, we used \texttt{GPT-4o} to generate a list of phrases that are characteristic of the ``aha moment'' (Appendix \ref{app:open-web-math-queries}), then used the MinHash algorithm \cite{666900} to search through  OpenWebMath \cite{paster2023openwebmathopendatasethighquality}, a dataset filtered from the CommonCrawl \cite{cc:Rana:2010:Common-Crawl-open-web-scale-crawl} frequently used in pre-training. We found that there was a significant number of matches in discussion forum threads, where the dialogue between multiple users showed similarity to long CoT, with multiple approaches being discussed along with backtracking and error correction (Appendix \ref{app:open-web-math-matches}). This raises the intriguing possibility that long CoT originated from human dialogue, although we should also note that discussion forums are a common source of data in OpenWebMath.

Based on these observations, we hypothesize that RL primarily guides the model to recombine skills it already internalized during pre-training towards new behaviors to improve performance on complex problem-solving tasks. Given the broad scope of this paper, we leave a more in-depth investigation of this behavior to future work.



\section{Discussions and Future Work}
In this work, we demystify long CoT reasoning in LLMs. In this section, we outline potential future directions.

\subsection{Scaling up Model Size}
We believe that model size is the primary factor limiting the emergence of the behavior observed in \autoref{result:base-reflections-existence}. Hyung Won Chung~\cite{Chung2024slides} recently shared a similar perspective, suggesting that smaller models may struggle to develop high-level reasoning skills and instead rely on heuristic-based pattern recognition. Future research could investigate RL using a larger base model.

\subsection{RL Infrastructure Is Still in Its Infancy
}
While attempting to scale up the model size, we encountered significant challenges in expanding to 32B, ultimately determining that the required number of GPUs was too large to proceed. We observe that open-source RL frameworks (e.g., OpenRLHF~\cite{hu2024openrlhfeasytousescalablehighperformance}) often coordinate multiple systems optimized for different training and inference workloads, leading to multiple copies of model parameters being stored in memory. Additionally, algorithms like PPO alternate between these workloads synchronously and sequentially, further limiting efficiency. These factors contribute to low hardware utilization, an issue that is particularly exacerbated in long CoT scenarios due to the higher variance in CoT length, which leads to stragglers during inference~\cite{kimi2025k15}. We look forward to advancements in machine learning and systems research that will help overcome these limitations and accelerate progress in long CoT modeling.

\subsection{REINFORCE Is More Tricky to Tune than PPO}
\label{result:reward-reinforce}

We also explored REINFORCE++~\cite{hu2025reinforce++} as a faster alternative to PPO for scaling up data. However, we found it to be significantly more unstable than PPO, leading to lower training accuracies (Figure \ref{fig:reinforce-instability}). As this instability may be due to an untuned setup (Appendix \ref{app:exp-hyperparams-reward-reinforce}), we refrain from making general claims about the algorithm. We present this as an observation that may be useful to the community.

\subsection{Scaling up Verification}

While our findings demonstrate that combining rule-based verifiers with prompt set filtering is highly effective, designing such rules and curating prompt sets across different domains remains labor-intensive. More fundamentally, this approach embeds human-designed heuristics into the RL environment, reflecting how we think rather than allowing for emergent learning. As highlighted in The Bitter Lesson\footnote{http://www.incompleteideas.net/IncIdeas/BitterLesson.html}, manually encoding human intuition tends to be an inefficient long-term strategy. This raises an intriguing question: how can verification signals be scaled effectively? Is there an equivalent of pretraining in the context of designing RL environments? We look forward to future research on silver supervision signals and the potential for self-supervised approaches in RL verification.

\subsection{Latent Capabilities in Base Models}

Reasoning is a latent capability in base models that has only recently been unlocked. Our analysis suggests that one possible source of this emergent thinking is human dialogue on Internet discussion forums. This raises a broader question: what other abilities exist, waiting to be elicited from the vast reservoir of human knowledge and experience embedded in pre-training data? We look forward to more detailed analyses tracing model behaviors back to their data origins, which could yield new insights and help uncover hidden capabilities within base models.

\section*{Impact Statement}

This paper aims to provide insights into scaling inference compute and training strategies to enable long chain-of-thought reasoning in large language models. The broader impacts of this work primarily relate to the potential for enhanced reasoning and problem-solving capabilities across various domains, where models capable of interpretable and verifiable reasoning could drive innovation and improve decision-making.
Our findings emphasize the importance of ensuring robust training data preparation, stability, and alignment with verifiable ground truths. We encourage future research to actively develop safeguards that ensure these capabilities are used responsibly. This includes careful design of reward shaping and training protocols to minimize unintended consequences while maximizing societal benefits.

\section*{Acknowledgment}
The authors would thank Yuanzhi Li for insightful discussions on this topic. The authors would also thank the SimpleRL team, particularly Weihao Zeng and Junxian He, for sharing their training experiences and experimental observations. Additionally, the authors appreciate Wenhu Chen, Xiaoyi Ren, Chao Li, Ziqiao Ma, Jiayi Pan, Xingyao Wang, and Seungone Kim for their valuable comments and discussions during the early or final stages of the project. Finally, the authors would acknowledge the DeepSeek-R1 and Kimi-k1.5 teams for their technical report releases, which inspired several additional experiment designs of this paper. This work was supported in part by a Carnegie Bosch Institute Fellowship to Xiang Yue.

\bibliography{reference}
\bibliographystyle{icml2025}

\newpage
\appendix
\onecolumn


\section{Related Work}
\paragraph{Complex reasoning and chain of thought prompting.} Large Language Models (LLMs) have demonstrated remarkable capabilities in various natural language processing tasks, including complex reasoning. A significant advancement in improving LLM reasoning ability is the implementation of Chain of Thought (CoT) prompting \cite{wei2022cot}. This technique involves guiding models to generate intermediate reasoning steps, thereby improving their performance on tasks that require logical deduction and multistep problem solving. Initial studies \cite{lambert2024tulu, wei2022cot, flan, yu2024metamath} focused on short CoT, where models produce concise reasoning paths to arrive at solutions. Although effective for straightforward problems, short CoT can be limiting when addressing more intricate tasks that necessitate deeper deliberation. OpenAI’s o1 \cite{openai2024o1} series models were the first to introduce inference-time scaling by increasing the length of the CoT reasoning process. This approach helps LLMs tackle complex problems by breaking them into finer steps and reflecting during problem-solving, leading to more accurate and comprehensive solutions. In this work, we explore long CoT by identifying key factors that enable models to exhibit this behavior, encouraging advanced reasoning capabilities.

\paragraph{Reinforcement learning for LLM.} Reinforcement Learning (RL) has proven effective in enhancing LLM performance across domains. RL techniques, such as Reinforcement Learning from Human Feedback (RLHF), align model outputs with human preferences, improving coherence \cite{ouyang2022training}. Recent studies \cite{kimi2025k15, deepseekai2025r1, lambert2024tulu} leverage RL to enable LLMs to explore reasoning paths autonomously for complex problems. DeepSeek-R1 \cite{deepseekai2025r1} achieves strong performance in mathematics, coding, and reasoning tasks without relying on a trained reward model \cite{lightman2023verifystep, wang2024multistep} or tree search \cite{feng2023alphazerolike, snell2024scaling}. Notably, this capability emerges even in base models without supervised fine-tuning, albeit at the cost of output readability. Similarly, Kimi K1.5 \cite{kimi2025k15} enhances general reasoning with RL, focusing on multimodal reasoning and controlling thought process length. These works highlight RL’s role in optimizing reasoning when intermediate steps are hard to supervise, and only final outcomes are verifiable. Our research share a similar setup but with more detail on disentangling how different model behaviors emerge under varying training conditions and initialization strategies.

\newpage
\section{Figures and Tables}

\begin{figure}[H]
    \centering
    \subfigure[Response lengths under different Cosine Rewards]
    {\includegraphics[width=0.24\textwidth]{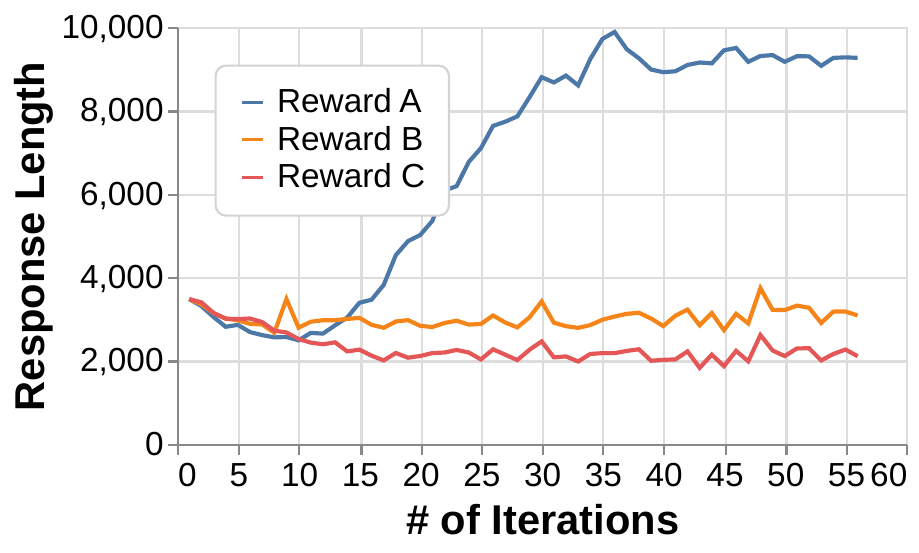}}
    \subfigure[Reward A]
    {\includegraphics[width=0.24\textwidth]{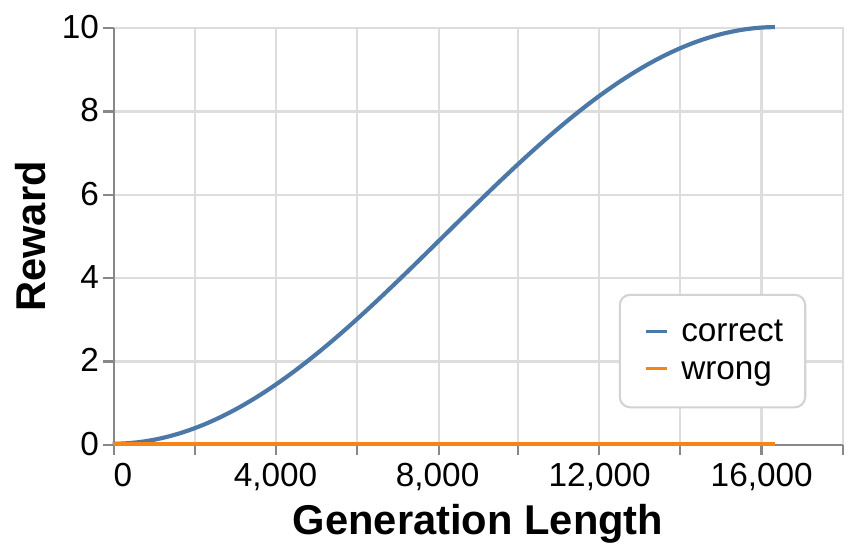}}
    \subfigure[Reward B]{\includegraphics[width=0.24\textwidth]{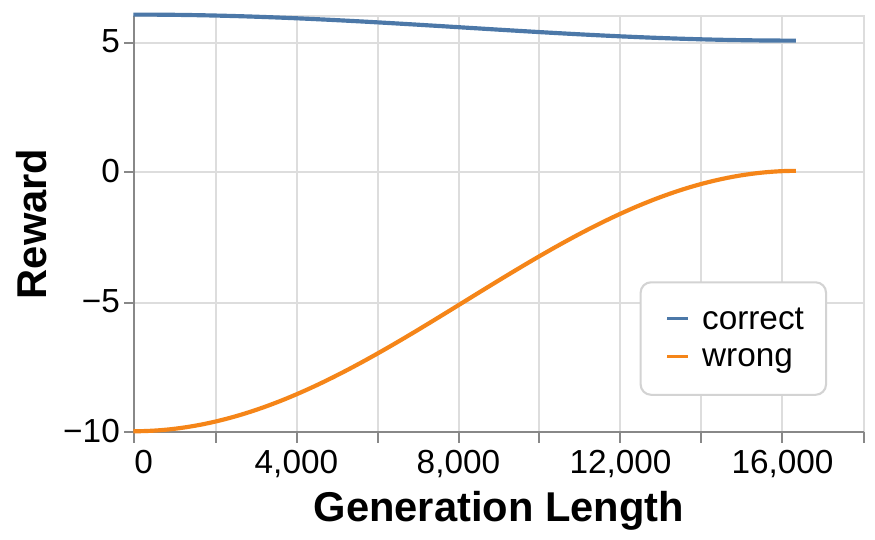}}
    \subfigure[Reward C]{\includegraphics[width=0.24\textwidth]{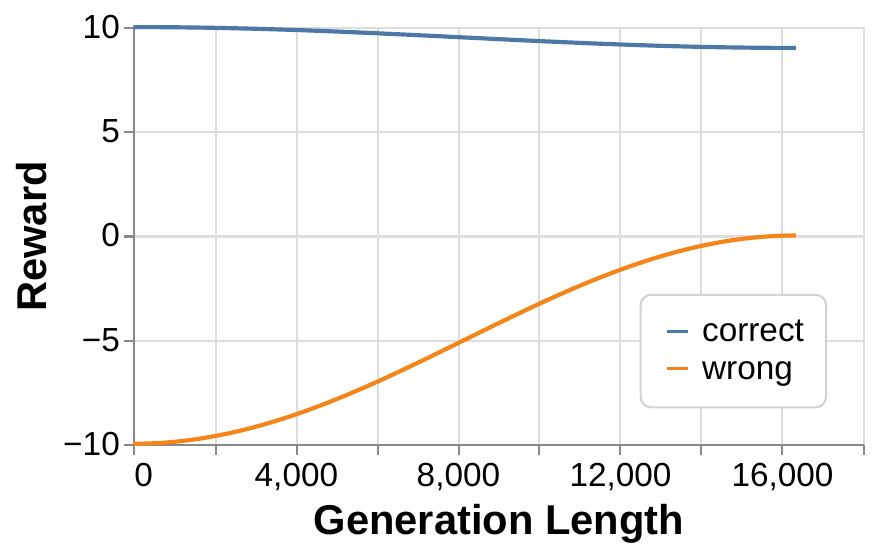}}
    \caption{(a) Tuning the hyperparameters of the Cosine Reward results in different length scaling behavior. Note that Reward A results in some performance degradation on downstream tasks due to the model's reduced ability to stop within the context window. (b) Reward A: $r_0^c=0, r_L^c=10, r_0^w=r_L^w=0$, (c) Reward B: $r_0^c = 6, r_L^c = 5, r_0^w = -10, r_L^w = 0$ (d) Reward C: $r_0^c = 10, r_L^c = 9, r_0^w = -10, r_L^w = 0$.}
    \label{fig:reward-cosine-hyperparams}
\end{figure}

\begin{figure}[H]
    \centering
    \includegraphics[width=0.6\linewidth]{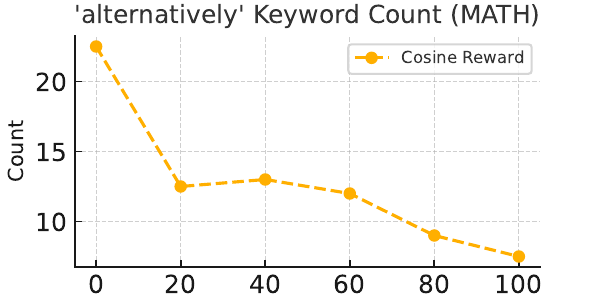}
    \vspace{-10pt}
    \caption{CoT branching frequency, estimated by the keyword count of the pivot word "alternatively,", decreased under the Cosine Reward with more training compute. We attributed this, along with increased repetition, to reward hacking.}
    \label{fig:reward-cosine-branching}
\end{figure}

\begin{figure}[H]
    \centering
    \includegraphics[width=0.5\linewidth]{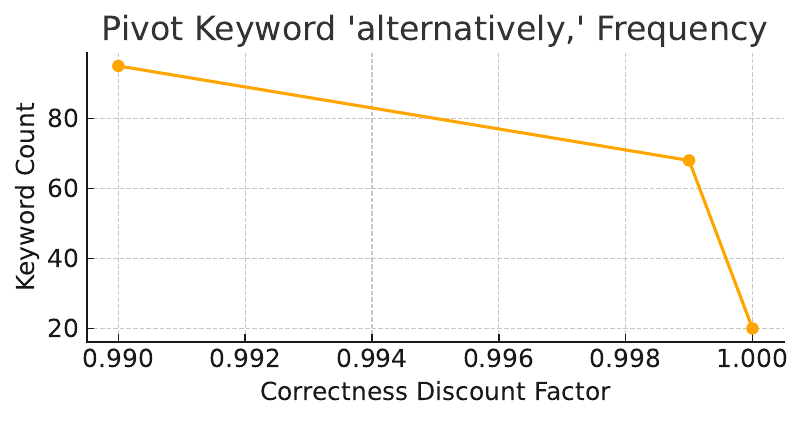}
    \vspace{-10pt}
   \caption{Branching frequency in CoT at different $\gamma_c$ values. Lowering the discount factor increased branching frequency, causing the model to abandon problem-solving approaches more quickly.}
    \label{fig:reward-indecisive}
\end{figure}

\begin{figure}[H]
    \centering
    \includegraphics[width=0.5\textwidth]{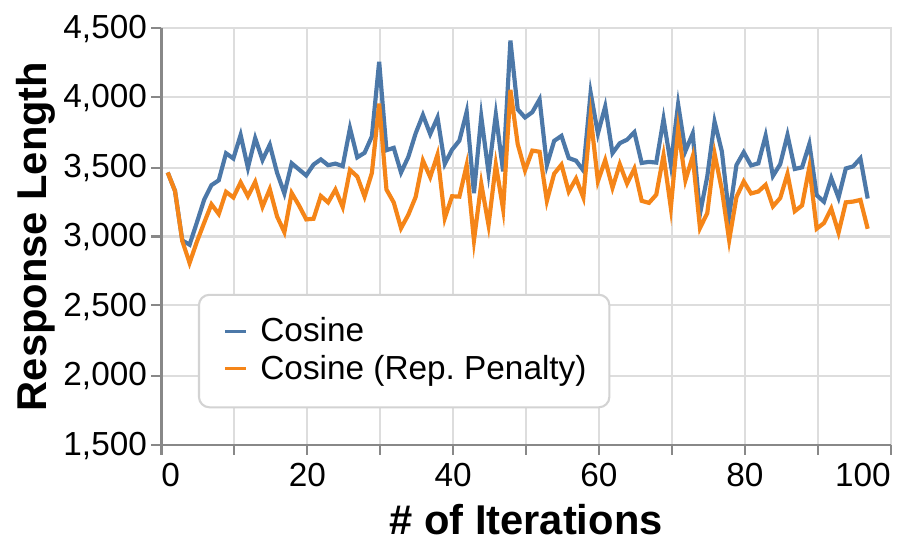}
    \caption{Training response length of models trained with Cosine Reward with and without repetition penalty. We see that repetition penalty reduced the length.}
    \label{fig:reward-repetition-penalty-length-effect}
\end{figure}

\begin{figure}[H]
    \centering
    \includegraphics[width=0.75\linewidth]{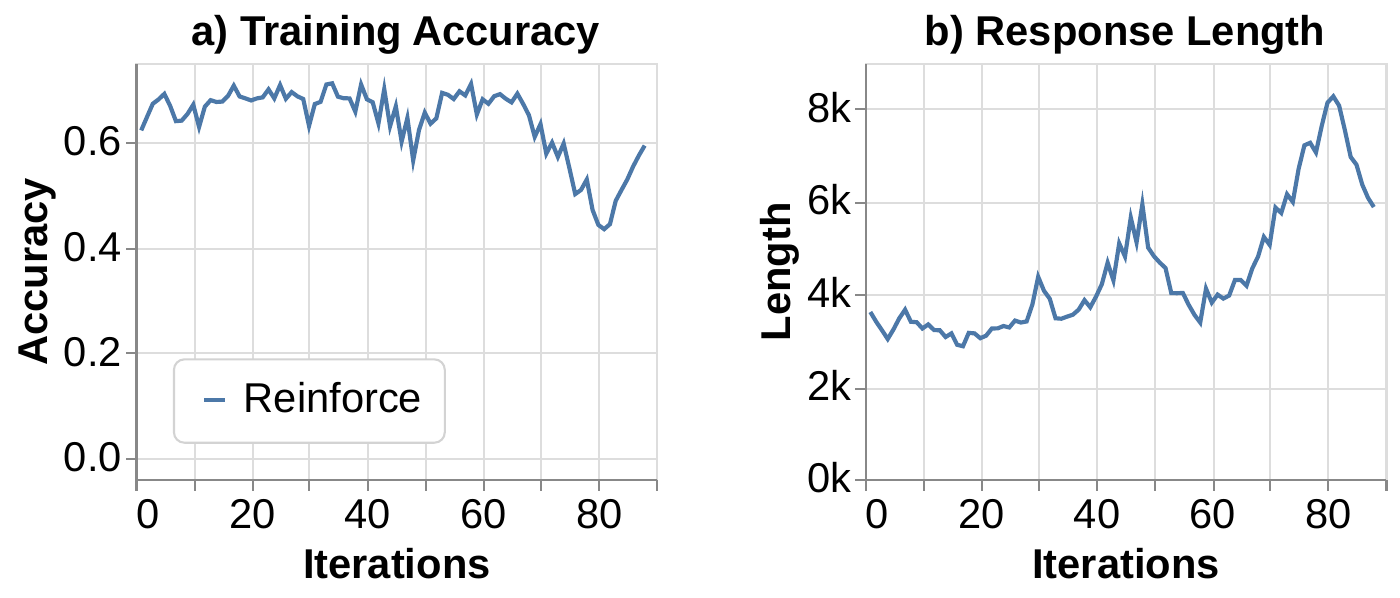}
    \caption{Reinforce with classic reward shows signs of training instability.}
    \label{fig:reinforce-instability}
\end{figure}


\begin{table}[H]
\small
\caption{Performance of model trained with different discount factors for the correctness (cosine) reward and repetition penalty. We see that different reward types have different optimal values.}
\vspace{10pt}
\centering
\begin{tabular}{@{}cccccc@{}}
\toprule
\begin{tabular}[c]{@{}c@{}}Correctness \\ Discount\end{tabular} & \begin{tabular}[c]{@{}c@{}}Repetition\\ Discount\end{tabular} & \begin{tabular}[c]{@{}c@{}}MATH\\ -500\end{tabular} & \begin{tabular}[c]{@{}c@{}}AIME \\ 2024\end{tabular} & \begin{tabular}[c]{@{}c@{}}Theo.\\ QA\end{tabular} & \begin{tabular}[c]{@{}c@{}}MMLU\\ -Pro-1k\end{tabular} \\ \midrule
\multicolumn{2}{c}{SFT} & 50.4 & 3.5 & 20.6 & 32.4 \\ \midrule
\multirow{3}{*}{1.000} & 1.000 & 55.7 & \textbf{5.0} & 25.7 & 34.5 \\
 & 0.999 & \textbf{58.0} & 4.6 & \textbf{26.0} & \textbf{36.5} \\
 & 0.99 & 57.8 & 3.8 & 24.5 & 33.3 \\ \midrule
\multirow{2}{*}{0.999} & 0.999 & 53.5 & 2.1 & 19.5 & 30.7 \\
 & 0.99 & 55.2 & 1.7 & 18.5 & 32.0 \\ \midrule
0.99 & 0.99 & 47.9 & 0.2 & 15.6 & 25.5 \\ \bottomrule
\end{tabular}%
\label{fig:multiple-gamma}
\end{table}

\newpage
\section{Algorithms and Formulas}

\subsection{Cosine Reward Formula}

\begin{equation}
\label{eqn:cosine-lr}
\textbf{CosFn}(t, T, \eta_{min}, \eta_{max}) = \eta_{min} + \frac{1}{2}(\eta_{max} - \eta_{min})(1 + \cos(\frac{t\pi}{T}))
\end{equation}

The formula above is commonly used as the learning rate schedule during gradient descent optimization. It was introduced by \cite{loshchilov2017sgdrstochasticgradientdescent}.

\subsection{N-gram Repetition Penalty}

\begin{algorithm}[H]
\caption{N-gram Repetition Penalty}\label{alg:reward-repetition-penalty}
\begin{algorithmic}[1]
    \STATE {\bfseries Input:} 
    \STATE \ \ \ \ $s$ : sequence of tokens
    \STATE \ \ \ \ $l$ : sequence length
    \STATE \ \ \ \ $N$ : n-gram size
    \STATE \ \ \ \ $P$ : penalty value
    \STATE \ \ \ \ $m$ : maximum sequence length
    \STATE {\bfseries Output:} $r \in \mathbb{R}^m$
    \STATE $seq \gets s[1:l]$ \COMMENT{Extract subsequence of length $l$}
    \STATE $ngrams \gets \emptyset$ \COMMENT{Set of observed n-grams}
    \STATE $r \gets \vec{0} \in \mathbb{R}^m$ \COMMENT{Initialize reward vector}
    \FOR{$j \gets 1$ to $|seq| - N + 1$}
        \STATE $ng \gets (seq[j], seq[j+1], ..., seq[j+N-1])$ \COMMENT{Current n-gram}
        \IF{$ng \in ngrams$}
            \FOR{$t \gets j$ to $j + N - 1$}
                \STATE $r[t] \gets P$ \COMMENT{Apply penalty}
            \ENDFOR
        \ENDIF
        \STATE $ngrams \gets ngrams \cup \{ng\}$
    \ENDFOR
    \STATE {\bfseries Output:} $r$
\end{algorithmic}
\end{algorithm}

\newpage
\section{Extracts}

\begin{tcolorbox}[label=extract:reward-short-term, colback=lightgray!10, colframe=black, title={Short-Term Thinking ($\gamma_c = 0.99$)}]
\textbf{But} in this case, we saw that the fixed points don't satisfy the equation 
\[
p(p(x)) = x.
\]

\textbf{Wait}, maybe the solutions to \( p(p(x)) = x \) are not the fixed points of \( p \), but something else.

\textbf{Alternatively}, perhaps I can consider that \( p(p(x)) = x \) implies that \( p \) is a bijection, 
but since \( p(x) \) is quadratic, it's not necessarily invertible over all real numbers.

\textbf{Wait}, but the problem says that \( p(p(x)) = x \) has four real solutions, so perhaps there are four 
values of \( x \) that satisfy this equation.

\textbf{Alternatively}, perhaps I need to find the roots of 
\[
p(p(x)) - x = 0,
\]
and solve for \( x \). But that seems complicated. Maybe there's a better way.
\end{tcolorbox}

\newpage
\section{Experimental Setup}

\subsection{Evaluation Setup} \label{app:eval-setup}

\begin{figure}[htbp]
    \centering
    \includegraphics[width=1\linewidth]{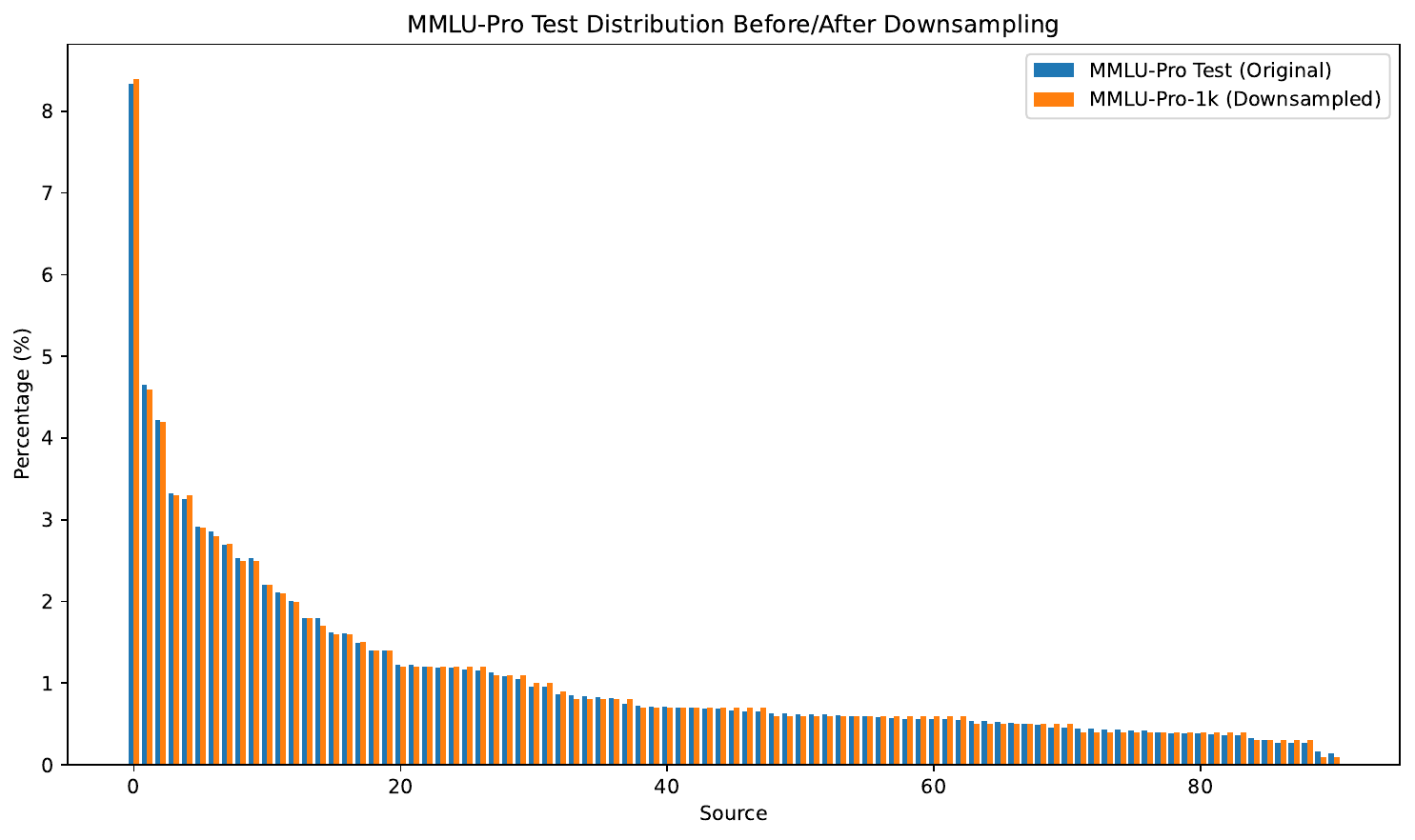}
    \vspace{-20pt}
    \caption{MMLU-Pro test distribution before/after downsampling for the MMLU-Pro-1k subset. The subset is i.i.d. to the full set.}
    \label{fig:mmlu-pro-test-downsample}
\end{figure}

\paragraph{Benchmarks} Below are details of our evaluation benchmarks:

\begin{itemize}
    \item \textbf{MATH-500} \citep{hendrycks2021math}: an in-domain mathematical reasoning benchmark. MATH consists of 12,500 problems from American high school math competitions. For efficiency, we adopt MATH-500, a widely-used i.i.d. subset of its test split.
    \item \textbf{AIME 2024}: an out-of-domain mathematical reasoning benchmark consisting of the 30 problems from American Invitational Mathematics Examination (AIME) 2024.
    \item \textbf{TheoremQA} \citep{chen2023theoremqa}: an out-of-domain STEM reasoning benchmark consisting of 800 samples. It covers 350+ theorems spanning across Math, EE\&CS, Physics and Finance.
    \item \textbf{MMLU-Pro-1k} \citep{wang2024mmlupro}: an out-of-domain general reasoning benchmark. MMLU-Pro comprises over 12,000 questions from academic exams and textbooks, spanning 14 diverse domains including Biology, Business, Chemistry, Computer Science, Economics, Engineering, Health, History, Law, Math, Philosophy, Physics, Psychology, and Others. For efficiency, we adopt an 1,000-sample i.i.d. subset of its test split, called MMLU-Pro-1k. We tried to keep the distribution identical to the original one. Figure \ref{fig:mmlu-pro-test-downsample} shows the distribution before/after the downsampling.
\end{itemize}

\paragraph{Statistical Metrics} We calculate the average accuracy with at least 4 random seeds. To tame the variance caused by the small size of AIME 2024, we sample 16 responses per prompt.

\paragraph{Implementation} We adopt the vLLM library to accelerate the inference and SymEval\footnote{\url{https://github.com/tongyx361/symeval}}, an elaborate answer grader capable of processing complex mathematical objects like matrices and functions, keeping consistent with the sampling and reward implementation in our RL setup. Note that a few RL experiments are carried out with an earlier version of the grader, causing nuanced performance differences.

\subsection{Details about Distillation}\label{app:distill}

To distill long CoT trajectories from \texttt{QwQ-32B-Preview}, we adopt the temperature $t=1.0$, the top-$p$ value of 0.95 and the maximum output length of 8192 tokens. Our preliminary experiments show that 8192 tokens show almost the same accuracy with \texttt{QwQ-32B-Preview} on MATH-500 as 16384 tokens, while costing significantly less time.

To distill short CoT trajectories from \texttt{Qwen2.5-Math-72B-Instruct}, we adopt the temperature $t=0.7$, the top-$p$ value of 0.95 and the maximum output length of 4096 tokens, since \texttt{Qwen2.5-Math-72B-Instruct} has a context limit of 4096 tokens and our preliminary experiments observe a non-negligible ratio of nonsense output when using $t=1.0$.

Note the data is distilled with SGLang \citep{zheng2024sglang} with an early version of our code.

When applying rejection sampling, we adopt the SymEval verifier as the grader.

\subsection{Details abour SFT Setup}\label{app:sft-setup}

We use OpenRLHF \citep{hu2024openrlhfeasytousescalablehighperformance} for our SFT experiments. By default, we adopt the SFT hyperparameters in Table \ref{tab:sft-hyperparams}.

For efficiency, we utilize Flash Attention 2 \citep{dao2023flashattention2} and ZeRO \citep{rajbhandari2020zero} based on the DeepSpeed library \citep{rasley2020deepspeed}. We uniformly set the micro batch size as 1 since we don't observe acceleration when increasing it.

\begin{table}[H]
\small
\caption{SFT Hyperparameters}
\centering
\begin{tabular}{@{}cccccc@{}}
\toprule
    \begin{tabular}[c]{@{}c@{}}Batch Size\end{tabular} &
    \begin{tabular}[c]{@{}c@{}}Context Length\end{tabular} &
    \begin{tabular}[c]{@{}c@{}}LR\end{tabular} &
    \begin{tabular}[c]{@{}c@{}}Epochs\end{tabular} \\
    \midrule
        256 &
        128K &
        5e-6 &
        2 \\
    \midrule
\end{tabular}
\label{tab:sft-hyperparams}
\end{table}

\subsection{Details about RL Setup}\label{app:rl-setup}

We use OpenRLHF \cite{hu2024openrlhfeasytousescalablehighperformance} for our RL experiments. When describing hyperparameters, we adopt the same naming conventions as OpenRLHF.

\subsection{Experiment Hyperparameters}\label{app:exp-hyperparams}

Note that the \texttt{BS} column below refers to both \texttt{rollout\_batch\_size} (the number of prompts used in a sampling-training iteration) and \texttt{train\_batch\_size} (the number of samples used in a training update) because we adopt the same number for these two hyperparameters in most of our RL setups. Also, the \texttt{Samples} column refers to the number of samples per prompt.

\subsubsection{Details of Section \ref{sec:sft-init-for-rl} (SFT Initialization for RL)}
\label{app:exp-hyperparams-sft-init-for-rl}

SFT Data: CoT data distilled from \texttt{QwQ-32B-Preview} or \texttt{Qwen2.5-Math-72B-Instruct} with the MATH train split with different number of candidate responses per prompt.

\begin{table}[H]
\small
\caption{Hyperparameters}
\vspace{10pt}
\centering
\begin{tabular}{@{}cccccccccc@{}}
\toprule
    \begin{tabular}[c]{@{}c@{}}Base Model\end{tabular} &
    \begin{tabular}[c]{@{}c@{}}Rewards\end{tabular} &
    \begin{tabular}[c]{@{}c@{}}GAE\end{tabular} &
    \begin{tabular}[c]{@{}c@{}}Episodes\end{tabular} &
    \begin{tabular}[c]{@{}c@{}}Samples\end{tabular} &
    \begin{tabular}[c]{@{}c@{}}BS\end{tabular} &
    \begin{tabular}[c]{@{}c@{}}Epochs\end{tabular} &
    \begin{tabular}[c]{@{}c@{}}Context Length\end{tabular} &
    \begin{tabular}[c]{@{}c@{}}LR\end{tabular} &
    \begin{tabular}[c]{@{}c@{}}KL\end{tabular} \\
    \midrule
        Llama3.1-8B &
        \makecell{
            Cosine: \\
            $r_0^c=+2$ \\
            $r_L^c=+1$ \\
            $r_0^w=-10$ \\
            $r_L^w=0$ \\
            $r_e=-10$ \\
            Rep. Penalty: \\
            $P=-0.05$ \\
            $N=40$} &
        \makecell{
            $\lambda=1$ \\
            $\gamma=1$} &
        4 & 4 & 512 & 1 &
        \makecell{Prompt: 2048 \\ Gen: 14336} &
        \makecell{Actor: 5e-7 \\ Critic: 9e-6} &
        0.01 \\
    \midrule
\end{tabular}
\label{tab:exp-hyperparams-sft-init-for-rl}
\end{table}

\subsubsection{Details of Section \ref{result:reward-length-stability} (CoT Length Stability)}
\label{app:exp-hyperparams-reward-length-stability}

SFT Data: Long CoT data distilled from \texttt{QwQ-32B-Preview} with the MATH train split.

\begin{table}[H]
\small
\caption{Hyperparameters}
\vspace{10pt}
\centering
\begin{tabular}{@{}cccccccccc@{}}
\toprule
    \begin{tabular}[c]{@{}c@{}}Base Model\end{tabular} &
    \begin{tabular}[c]{@{}c@{}}Rewards\end{tabular} &
    \begin{tabular}[c]{@{}c@{}}GAE\end{tabular} &
    \begin{tabular}[c]{@{}c@{}}Episodes\end{tabular} &
    \begin{tabular}[c]{@{}c@{}}Samples\end{tabular} &
    \begin{tabular}[c]{@{}c@{}}BS\end{tabular} &
    \begin{tabular}[c]{@{}c@{}}Epochs\end{tabular} &
    \begin{tabular}[c]{@{}c@{}}Context Length\end{tabular} &
    \begin{tabular}[c]{@{}c@{}}LR\end{tabular} &
    \begin{tabular}[c]{@{}c@{}}KL\end{tabular} \\
    \midrule
        Llama3.1-8B &
        Correct: $+1$ &
        \makecell{
            $\lambda=1$ \\
            $\gamma=1$} &
        8 & 8 & 512 & 1 &
        \makecell{Prompt: 2048 \\ Gen: 14336} &
        \makecell{Actor: 5e-7 \\ Critic: 4.5e-6} &
        0.01 \\
    \midrule
        Qwen2.5-Math-7B &
        Correct: $+1$ &
        \makecell{
            $\lambda=1$ \\
            $\gamma=1$} &
        8 & 8 & 512 & 1 &
        \makecell{Prompt: 2048 \\ Gen: 14336} &
        \makecell{Actor: 5e-7 \\ Critic: 4.5e-6} &
        0.01 \\
    \midrule
\end{tabular}
\label{fig:exp-hyperparams-reward-length-stability}
\end{table}

\newpage
\subsubsection{Details of Section \ref{result:reward-length-scaling} (Active Scaling of CoT Length)}
\label{app:exp-hyperparams-reward-length-scaling}

SFT Data: Long CoT data distilled from \texttt{QwQ-32B-Preview} with the MATH train split.

\begin{table}[H]
\small
\caption{Hyperparameters}
\vspace{10pt}
\centering
\begin{tabular}{@{}cccccccccc@{}}
\toprule
    \begin{tabular}[c]{@{}c@{}}Base Model\end{tabular} &
    \begin{tabular}[c]{@{}c@{}}Rewards\end{tabular} &
    \begin{tabular}[c]{@{}c@{}}GAE\end{tabular} &
    \begin{tabular}[c]{@{}c@{}}Episodes\end{tabular} &
    \begin{tabular}[c]{@{}c@{}}Samples\end{tabular} &
    \begin{tabular}[c]{@{}c@{}}BS\end{tabular} &
    \begin{tabular}[c]{@{}c@{}}Epochs\end{tabular} &
    \begin{tabular}[c]{@{}c@{}}Context Length\end{tabular} &
    \begin{tabular}[c]{@{}c@{}}LR\end{tabular} &
    \begin{tabular}[c]{@{}c@{}}KL\end{tabular} \\
    \midrule
        Llama3.1-8B &
        Correct: $+1$ &
        \makecell{
            $\lambda=1$ \\
            $\gamma=1$} &
        8 & 8 & 512 & 1 &
        \makecell{Prompt: 2048 \\ Gen: 14336} &
        \makecell{Actor: 5e-7 \\ Critic: 4.5e-6} &
        0.01 \\
    \midrule
        Llama3.1-8B &
        \makecell{Cosine: \\ $r_0^c=+2$ \\ $r_L^c=+1$ \\ $r_0^w=-10$ \\ $r_L^w=0$ \\ $r_e=-10$} &
        \makecell{
            $\lambda=1$ \\
            $\gamma=1$} &
        8 & 8 & 512 & 1 &
        \makecell{Prompt: 2048 \\ Gen: 14336} &
        \makecell{Actor: 5e-7 \\ Critic: 4.5e-6} &
        0.01 \\
    \midrule
        Llama3.1-8B &
        Correct: $+1$ &
        \makecell{
            $\lambda=1$ \\
            $\gamma=1$} &
        8 & 16 & 512 & 2 &
        \makecell{Prompt: 2048 \\ Gen: 14336} &
        \makecell{Actor: 5e-7 \\ Critic: 9e-6} &
        0.01 \\
    \midrule
        Llama3.1-8B &
        \makecell{
            Cosine: \\
            $r_0^c=+2$ \\
            $r_L^c=+1$ \\
            $r_0^w=-10$ \\
            $r_L^w=0$ \\
            $r_e=-10$} &
        \makecell{
            $\lambda=1$ \\
            $\gamma=1$} &
        8 & 16 & 512 & 2 &
        \makecell{Prompt: 2048 \\ Gen: 14336} &
        \makecell{Actor: 5e-7 \\ Critic: 9e-6} &
        0.01 \\
    \midrule
        Llama3.1-8B &
        \makecell{
            Cosine: \\
            $r_0^c=+2$ \\
            $r_L^c=+1$ \\
            $r_0^w=-10$ \\
            $r_L^w=0$ \\
            $r_e=-10$ \\
            Rep. Penalty: \\
            $P=-0.05$ \\
            $N=40$} &
        \makecell{
            $\lambda=1$ \\
            $\gamma_c=1$ \\
            $\gamma_p=0.99$} &
        8 & 16 & 512 & 2 &
        \makecell{Prompt: 2048 \\ Gen: 14336} &
        \makecell{Actor: 5e-7 \\ Critic: 9e-6} &
        0.01 \\
    \midrule
\end{tabular}
\label{fig:exp-hyperparams-reward-length-scaling}
\end{table}

\newpage
\subsubsection{Details of Section \ref{result:reward-cosine-hyperparams} (Cosine Reward Hyperparameters)}
\label{app:exp-hyperparams-reward-cosine-hyperparams}

SFT Data: Long CoT data distilled from \texttt{QwQ-32B-Preview} with the MATH train split.

\begin{table}[H]
\small
\caption{Hyperparameters}
\vspace{10pt}
\centering
\begin{tabular}{@{}cccccccccc@{}}
\toprule
    \begin{tabular}[c]{@{}c@{}}Base Model\end{tabular} &
    \begin{tabular}[c]{@{}c@{}}Rewards\end{tabular} &
    \begin{tabular}[c]{@{}c@{}}GAE\end{tabular} &
    \begin{tabular}[c]{@{}c@{}}Episodes\end{tabular} &
    \begin{tabular}[c]{@{}c@{}}Samples\end{tabular} &
    \begin{tabular}[c]{@{}c@{}}BS\end{tabular} &
    \begin{tabular}[c]{@{}c@{}}Epochs\end{tabular} &
    \begin{tabular}[c]{@{}c@{}}Context Length\end{tabular} &
    \begin{tabular}[c]{@{}c@{}}LR\end{tabular} &
    \begin{tabular}[c]{@{}c@{}}KL\end{tabular} \\
    \midrule
        Llama3.1-8B &
        \makecell{
            Cosine: \\
            $r_0^c=0$ \\
            $r_L^c=+10$ \\
            $r_0^w=0$ \\
            $r_L^w=0$ \\
            $r_e=-10$ \\
            Rep. Penalty: \\
            $P=-0.05$ \\
            $N=40$} &
        \makecell{
            $\lambda=1$ \\
            $\gamma_c=1$ \\
            $\gamma_p=0.99$} &
        4 & 4 & 512 & 1 &
        \makecell{Prompt: 2048 \\ Gen: 14336} &
        \makecell{Actor: 5e-7 \\ Critic: 9e-6} &
        0.01 \\
    \midrule
        Llama3.1-8B &
        \makecell{
            Cosine: \\
            $r_0^c=+6$ \\
            $r_L^c=+5$ \\
            $r_0^w=-10$ \\
            $r_L^w=0$ \\
            $r_e=-10$ \\
            Rep. Penalty: \\
            $P=-0.05$ \\
            $N=40$} &
        \makecell{
            $\lambda=1$ \\
            $\gamma_c=1$ \\
            $\gamma_p=0.99$} &
        4 & 4 & 512 & 1 &
        \makecell{Prompt: 2048 \\ Gen: 14336} &
        \makecell{Actor: 5e-7 \\ Critic: 9e-6} &
        0.01 \\
    \midrule
        Llama3.1-8B &
        \makecell{
            Cosine: \\
            $r_0^c=+10$ \\
            $r_L^c=+9$ \\
            $r_0^w=-10$ \\
            $r_L^w=0$ \\
            $r_e=-10$ \\
            Rep. Penalty: \\
            $P=-0.05$ \\
            $N=40$} &
        \makecell{
            $\lambda=1$ \\
            $\gamma_c=1$ \\
            $\gamma_p=0.99$} &
        4 & 4 & 512 & 1 &
        \makecell{Prompt: 2048 \\ Gen: 14336} &
        \makecell{Actor: 5e-7 \\ Critic: 9e-6} &
        0.01 \\
    \midrule
\end{tabular}
\label{fig:exp-hyperparams-reward-cosine-hyperparams}
\end{table}

\newpage
\subsubsection{Details of Section \ref{result:reward-context-window} (Context Window Size)}
\label{app:exp-hyperparams-reward-context-window}

SFT Data: Long CoT data distilled from \texttt{QwQ-32B-Preview} with the MATH train split.

\begin{table}[H]
\small
\caption{Hyperparameters}
\vspace{10pt}
\centering
\begin{tabular}{@{}cccccccccc@{}}
\toprule
    \begin{tabular}[c]{@{}c@{}}Base Model\end{tabular} &
    \begin{tabular}[c]{@{}c@{}}Rewards\end{tabular} &
    \begin{tabular}[c]{@{}c@{}}GAE\end{tabular} &
    \begin{tabular}[c]{@{}c@{}}Episodes\end{tabular} &
    \begin{tabular}[c]{@{}c@{}}Samples\end{tabular} &
    \begin{tabular}[c]{@{}c@{}}BS\end{tabular} &
    \begin{tabular}[c]{@{}c@{}}Epochs\end{tabular} &
    \begin{tabular}[c]{@{}c@{}}Context Length\end{tabular} &
    \begin{tabular}[c]{@{}c@{}}LR\end{tabular} &
    \begin{tabular}[c]{@{}c@{}}KL\end{tabular} \\
    \midrule
        Llama3.1-8B &
        \makecell{
            Cosine: \\
            $r_0^c=+2$ \\
            $r_L^c=+1$ \\
            $r_0^w=-10$ \\
            $r_L^w=0$ \\
            $r_e=-10$ \\
            Rep. Penalty: \\
            $P=-0.05$ \\
            $N=40$} &
        \makecell{
            $\lambda=1$ \\
            $\gamma_c=1$ \\
            $\gamma_p=0.99$} &
        8 & 8 & 512 & 1 &
        \makecell{Prompt: 2048 \\ Gen: 2048} &
        \makecell{Actor: 5e-7 \\ Critic: 9e-6} &
        0.01 \\
    \midrule
        Llama3.1-8B &
        \makecell{
            Cosine: \\
            $r_0^c=+2$ \\
            $r_L^c=+1$ \\
            $r_0^w=-10$ \\
            $r_L^w=0$ \\
            $r_e=-10$ \\
            Rep. Penalty: \\
            $P=-0.05$ \\
            $N=40$} &
        \makecell{
            $\lambda=1$ \\
            $\gamma_c=1$ \\
            $\gamma_p=0.99$} &
        8 & 8 & 512 & 1 &
        \makecell{Prompt: 2048 \\ Gen: 6144} &
        \makecell{Actor: 5e-7 \\ Critic: 9e-6} &
        0.01 \\
    \midrule
        Llama3.1-8B &
        \makecell{
            Cosine: \\
            $r_0^c=+2$ \\
            $r_L^c=+1$ \\
            $r_0^w=-10$ \\
            $r_L^w=0$ \\
            $r_e=-10$ \\
            Rep. Penalty: \\
            $P=-0.05$ \\
            $N=40$} &
        \makecell{
            $\lambda=1$ \\
            $\gamma_c=1$ \\
            $\gamma_p=0.99$} &
        8 & 8 & 512 & 1 &
        \makecell{Prompt: 2048 \\ Gen: 14336} &
        \makecell{Actor: 5e-7 \\ Critic: 9e-6} &
        0.01 \\
    \midrule
\end{tabular}
\label{fig:exp-hyperparams-reward-context-window}
\end{table}

\subsubsection{Details of Section \ref{result:reward-hacking} (Length Reward Hacking)}
\label{app:exp-hyperparams-reward-hacking}

SFT Data: Long CoT data distilled from \texttt{QwQ-32B-Preview} with the MATH train split.

\begin{table}[H]
\small
\caption{Hyperparameters}
\vspace{10pt}
\centering
\begin{tabular}{@{}cccccccccc@{}}
\toprule
    \begin{tabular}[c]{@{}c@{}}Base Model\end{tabular} &
    \begin{tabular}[c]{@{}c@{}}Rewards\end{tabular} &
    \begin{tabular}[c]{@{}c@{}}GAE\end{tabular} &
    \begin{tabular}[c]{@{}c@{}}Episodes\end{tabular} &
    \begin{tabular}[c]{@{}c@{}}Samples\end{tabular} &
    \begin{tabular}[c]{@{}c@{}}BS\end{tabular} &
    \begin{tabular}[c]{@{}c@{}}Epochs\end{tabular} &
    \begin{tabular}[c]{@{}c@{}}Context Length\end{tabular} &
    \begin{tabular}[c]{@{}c@{}}LR\end{tabular} &
    \begin{tabular}[c]{@{}c@{}}KL\end{tabular} \\
    \midrule
        Llama3.1-8B &
        \makecell{
            Cosine: \\
            $r_0^c=+2$ \\
            $r_L^c=+1$ \\
            $r_0^w=-10$ \\
            $r_L^w=0$ \\
            $r_e=-10$} &
        \makecell{
            $\lambda=1$ \\
            $\gamma=1$} &
        8 & 16 & 512 & 2 &
        \makecell{Prompt: 2048 \\ Gen: 14336} &
        \makecell{Actor: 5e-7 \\ Critic: 9e-6} &
        0.01 \\
    \midrule
        Llama3.1-8B &
        \makecell{
            Cosine: \\
            $r_0^c=+2$ \\
            $r_L^c=+1$ \\
            $r_0^w=-10$ \\
            $r_L^w=0$ \\
            $r_e=-10$ \\
            Rep. Penalty: \\
            $P=-0.05$ \\
            $N=40$} &
        \makecell{
            $\lambda=1$ \\
            $\gamma_c=1$ \\
            $\gamma_p=0.99$} &
        8 & 16 & 512 & 2 &
        \makecell{Prompt: 2048 \\ Gen: 14336} &
        \makecell{Actor: 5e-7 \\ Critic: 9e-6} &
        0.01 \\
    \midrule
\end{tabular}
\label{fig:exp-hyperparams-reward-hacking}
\end{table}

\subsubsection{Details of Section \ref{result:optimal-discount} (Optimal Discount Factors)}
\label{app:exp-hyperparams-optimal-discount}

SFT Data: Long CoT data distilled from \texttt{QwQ-32B-Preview} with the MATH train split.

\begin{table}[H]
\small
\caption{Hyperparameters}
\centering
\begin{tabular}{@{}cccccccccc@{}}
\toprule
    \begin{tabular}[c]{@{}c@{}}Base Model\end{tabular} &
    \begin{tabular}[c]{@{}c@{}}Rewards\end{tabular} &
    \begin{tabular}[c]{@{}c@{}}GAE\end{tabular} &
    \begin{tabular}[c]{@{}c@{}}Episodes\end{tabular} &
    \begin{tabular}[c]{@{}c@{}}Samples\end{tabular} &
    \begin{tabular}[c]{@{}c@{}}BS\end{tabular} &
    \begin{tabular}[c]{@{}c@{}}Epochs\end{tabular} &
    \begin{tabular}[c]{@{}c@{}}Context Length\end{tabular} &
    \begin{tabular}[c]{@{}c@{}}LR\end{tabular} &
    \begin{tabular}[c]{@{}c@{}}KL\end{tabular} \\
    \midrule
        Llama3.1-8B &
        \makecell{
            Cosine: \\
            $r_0^c=+2$ \\
            $r_L^c=+1$ \\
            $r_0^w=-10$ \\
            $r_L^w=0$ \\
            $r_e=-10$ \\
            Rep. Penalty: \\
            $P=-0.05$ \\
            $N=40$} &
        \makecell{
            $\lambda=1$ \\
            $\gamma_c=1$ \\
            $\gamma_p=1$} &
        4 & 4 & 512 & 1 &
        \makecell{Prompt: 2048 \\ Gen: 14336} &
        \makecell{Actor: 5e-7 \\ Critic: 9e-6} &
        0.01 \\
    \midrule
        Llama3.1-8B &
        \makecell{
            Cosine: \\
            $r_0^c=+2$ \\
            $r_L^c=+1$ \\
            $r_0^w=-10$ \\
            $r_L^w=0$ \\
            $r_e=-10$ \\
            Rep. Penalty: \\
            $P=-0.05$ \\
            $N=40$} &
        \makecell{
            $\lambda=1$ \\
            $\gamma_c=1$ \\
            $\gamma_p=0.999$} &
        4 & 4 & 512 & 1 &
        \makecell{Prompt: 2048 \\ Gen: 14336} &
        \makecell{Actor: 5e-7 \\ Critic: 9e-6} &
        0.01 \\
    \midrule
        Llama3.1-8B &
        \makecell{
            Cosine: \\
            $r_0^c=+2$ \\
            $r_L^c=+1$ \\
            $r_0^w=-10$ \\
            $r_L^w=0$ \\
            $r_e=-10$ \\
            Rep. Penalty: \\
            $P=-0.05$ \\
            $N=40$} &
        \makecell{
            $\lambda=1$ \\
            $\gamma_c=1$ \\
            $\gamma_p=0.99$} &
        4 & 4 & 512 & 1 &
        \makecell{Prompt: 2048 \\ Gen: 14336} &
        \makecell{Actor: 5e-7 \\ Critic: 9e-6} &
        0.01 \\
    \midrule
        Llama3.1-8B &
        \makecell{
            Cosine: \\
            $r_0^c=+2$ \\
            $r_L^c=+1$ \\
            $r_0^w=-10$ \\
            $r_L^w=0$ \\
            $r_e=-10$ \\
            Rep. Penalty: \\
            $P=-0.05$ \\
            $N=40$} &
        \makecell{
            $\lambda=1$ \\
            $\gamma_c=0.999$ \\
            $\gamma_p=0.999$} &
        4 & 4 & 512 & 1 &
        \makecell{Prompt: 2048 \\ Gen: 14336} &
        \makecell{Actor: 5e-7 \\ Critic: 9e-6} &
        0.01 \\
    \midrule
        Llama3.1-8B &
        \makecell{
            Cosine: \\
            $r_0^c=+2$ \\
            $r_L^c=+1$ \\
            $r_0^w=-10$ \\
            $r_L^w=0$ \\
            $r_e=-10$ \\
            Rep. Penalty: \\
            $P=-0.05$ \\
            $N=40$} &
        \makecell{
            $\lambda=1$ \\
            $\gamma_c=0.999$ \\
            $\gamma_p=0.99$} &
        4 & 4 & 512 & 1 &
        \makecell{Prompt: 2048 \\ Gen: 14336} &
        \makecell{Actor: 5e-7 \\ Critic: 9e-6} &
        0.01 \\
    \midrule
        Llama3.1-8B &
        \makecell{
            Cosine: \\
            $r_0^c=+2$ \\
            $r_L^c=+1$ \\
            $r_0^w=-10$ \\
            $r_L^w=0$ \\
            $r_e=-10$ \\
            Rep. Penalty: \\
            $P=-0.05$ \\
            $N=40$} &
        \makecell{
            $\lambda=1$ \\
            $\gamma_c=0.99$ \\
            $\gamma_p=0.99$} &
        4 & 4 & 512 & 1 &
        \makecell{Prompt: 2048 \\ Gen: 14336} &
        \makecell{Actor: 5e-7 \\ Critic: 9e-6} &
        0.01 \\
    \midrule
\end{tabular}
\label{fig:exp-hyperparams-optimal-discount}
\end{table}

\subsubsection{Details of Section \ref{result:reward-verify-clean} (RL with Noisy Verifiable Data)}
\label{app:exp-hyperparams-reward-verify-clean}

SFT Data: 115k filtered from 462k instances of long CoT data distilled from \texttt{QwQ-32B-Preview} with WebInstruct.

\begin{table}[H]
\small
\caption{Hyperparameters}
\vspace{10pt}
\centering
\begin{tabular}{@{}cccccccccc@{}}
\toprule
    \begin{tabular}[c]{@{}c@{}}Base Model\end{tabular} &
    \begin{tabular}[c]{@{}c@{}}RL Prompt Set \\ Verifier\end{tabular} &
    \begin{tabular}[c]{@{}c@{}}Rewards\end{tabular} &
    \begin{tabular}[c]{@{}c@{}}GAE\end{tabular} &
    \begin{tabular}[c]{@{}c@{}}Episodes \\ Instances\end{tabular} &
    \begin{tabular}[c]{@{}c@{}}Samples\end{tabular} &
    \begin{tabular}[c]{@{}c@{}}BS\end{tabular} &
    \begin{tabular}[c]{@{}c@{}}Epochs\end{tabular} &
    \begin{tabular}[c]{@{}c@{}}Context Length\end{tabular} &
    \begin{tabular}[c]{@{}c@{}}LR \\ KL\end{tabular}\\
    \midrule
        Llama3.1-8B &
        \makecell{
            Unfiltered \\
            (30k sampled) \\
            Symeval} &
        \makecell{
            Cosine: \\
            $r_0^c=+2$ \\
            $r_L^c=+1$ \\
            $r_0^w=-10$ \\
            $r_L^w=0$ \\
            $r_e=-10$ \\
            Rep. Penalty: \\
            $P=-0.05$ \\
            $N=40$} &
        \makecell{
            $\lambda=1$ \\
            $\gamma_c=1$ \\
            $\gamma_p=0.99$} &
        \makecell{
            1 \\
            30k instances} &
        4 & 512 & 1 &
        \makecell{Prompt: 2048 \\ Gen: 14336} &
        \makecell{Actor: 5e-7 \\ Critic: 9e-6 \\
        KL: 0.01} \\
    \midrule
        Llama3.1-8B &
        \makecell{
            Unfiltered \\
            (30k sampled) \\
            LLM-as-a-judge} &
        \makecell{
            Cosine: \\
            $r_0^c=+2$ \\
            $r_L^c=+1$ \\
            $r_0^w=-10$ \\
            $r_L^w=0$ \\
            $r_e=-10$ \\
            Rep. Penalty: \\
            $P=-0.05$ \\
            $N=40$} &
        \makecell{
            $\lambda=1$ \\
            $\gamma_c=1$ \\
            $\gamma_p=0.99$} &
        \makecell{
            1 \\
            30k instances} &
        4 & 512 & 1 &
        \makecell{Prompt: 2048 \\ Gen: 14336} &
        \makecell{Actor: 5e-7 \\ Critic: 9e-6 \\
        KL: 0.01} \\
    \midrule
        Llama3.1-8B &
        \makecell{
            Filtered \\
            (30k sampled) \\
            Symeval} &
        \makecell{
            Cosine: \\
            $r_0^c=+2$ \\
            $r_L^c=+1$ \\
            $r_0^w=-10$ \\
            $r_L^w=0$ \\
            $r_e=-10$ \\
            Rep. Penalty: \\
            $P=-0.05$ \\
            $N=40$} &
        \makecell{
            $\lambda=1$ \\
            $\gamma_c=1$ \\
            $\gamma_p=0.99$} &
        \makecell{
            1 \\
            30k instances} &
        4 & 512 & 1 &
        \makecell{Prompt: 2048 \\ Gen: 14336} &
        \makecell{Actor: 5e-7 \\ Critic: 9e-6 \\
        KL: 0.01} \\
    \midrule
        Llama3.1-8B &
        \makecell{
            Filtered \\
            (30k sampled) \\
            LLM-as-a-judge} &
        \makecell{
            Cosine: \\
            $r_0^c=+2$ \\
            $r_L^c=+1$ \\
            $r_0^w=-10$ \\
            $r_L^w=0$ \\
            $r_e=-10$ \\
            Rep. Penalty: \\
            $P=-0.05$ \\
            $N=40$} &
        \makecell{
            $\lambda=1$ \\
            $\gamma_c=1$ \\
            $\gamma_p=0.99$} &
        \makecell{
            1 \\
            30k instances} &
        4 & 512 & 1 &
        \makecell{Prompt: 2048 \\ Gen: 14336} &
        \makecell{Actor: 5e-7 \\ Critic: 9e-6 \\
        KL: 0.01} \\
    \midrule
\end{tabular}
\label{fig:exp-hyperparams-reward-verify-clean}
\end{table}

\subsubsection{Details of Section \ref{sec:rl-from-base} (Exploration on RL from the Base Model)}
\label{app:exp-hyperparams-rl-from-base}

\begin{table}[H]
\small
\caption{Hyperparameters}
\vspace{10pt}
\centering
\begin{tabular}{@{}cccccccccc@{}}
\toprule
    \begin{tabular}[c]{@{}c@{}}Base Model\end{tabular} &
    \begin{tabular}[c]{@{}c@{}}Rewards\end{tabular} &
    \begin{tabular}[c]{@{}c@{}}GAE\end{tabular} &
    \begin{tabular}[c]{@{}c@{}}Episodes\end{tabular} &
    \begin{tabular}[c]{@{}c@{}}Samples\end{tabular} &
    \begin{tabular}[c]{@{}c@{}}BS\end{tabular} &
    \begin{tabular}[c]{@{}c@{}}Epochs\end{tabular} &
    \begin{tabular}[c]{@{}c@{}}Context Length\end{tabular} &
    \begin{tabular}[c]{@{}c@{}}LR\end{tabular} &
    \begin{tabular}[c]{@{}c@{}}KL\end{tabular} \\
    \midrule
        Qwen2.5-Math-7B &
        \makecell{
            Correct: $+1$ \\
            Wrong: $-0.5$ \\
            No Answer: $-1$} &
        \makecell{
            $\lambda=0.95$ \\
            $\gamma=1$} &
        20 & 8 & 
        \makecell{
            1024 \\
            (Train: 128)
        }
        & 1 &
        \makecell{Prompt: 1024 \\ Gen: 3072} &
        \makecell{Actor: 5e-7 \\ Critic: 9e-6} &
        0.01 \\
    \midrule
        Qwen2.5-Math-7B &
        Correct: $+1$ &
        \makecell{
            $\lambda=1$ \\
            $\gamma=1$} &
        8 & 8 & 512 & 1 &
        \makecell{Prompt: 2048 \\ Gen: 14336 } &
        \makecell{Actor: 5e-7 \\ Critic: 4.5e-6} &
        0.01 \\
    \midrule
        Qwen2.5-Math-7B &
        \makecell{
            Cosine: \\
            $r_0^c=+2$ \\
            $r_L^c=+1$ \\
            $r_0^w=-10$ \\
            $r_L^w=0$ \\
            $r_e=-10$ \\
            Rep. Penalty: \\
            $P=-0.05$ \\
            $N=40$} &
        \makecell{
            $\lambda=1$ \\
            $\gamma=1$} &
        8 & 8 & 512 & 1 &
        \makecell{Prompt: 2048 \\ Gen: 14336} &
        \makecell{Actor: 5e-7 \\ Critic: 4.5e-6} &
        0.01 \\
    \midrule
\end{tabular}
\label{tab:exp-hyperparams-rl-from-base}
\end{table}

\subsubsection{Details of Section \ref{result:reward-reinforce} (REINFORCE is more tricky to tune than PPO)}
\label{app:exp-hyperparams-reward-reinforce}

SFT Data: Long CoT data distilled from \texttt{QwQ-32B-Preview} with the MATH train split.

\begin{table}[H]
\small
\caption{Hyperparameters}
\vspace{10pt}
\centering
\begin{tabular}{@{}ccccccccccc@{}}
\toprule
    \begin{tabular}[c]{@{}c@{}}Base Model\end{tabular} &
    \begin{tabular}[c]{@{}c@{}}Rewards\end{tabular} &
    \begin{tabular}[c]{@{}c@{}}Gamma\end{tabular} &
    \begin{tabular}[c]{@{}c@{}}Episodes\end{tabular} &
    \begin{tabular}[c]{@{}c@{}}Samples\end{tabular} &
    \begin{tabular}[c]{@{}c@{}}BS\end{tabular} &
    \begin{tabular}[c]{@{}c@{}}Epochs\end{tabular} &
    \begin{tabular}[c]{@{}c@{}}Context Length\end{tabular} &
    \begin{tabular}[c]{@{}c@{}}LR\end{tabular} &
    \begin{tabular}[c]{@{}c@{}}KL\end{tabular} &
    \begin{tabular}[c]{@{}c@{}}Clip\end{tabular} \\
    \midrule
        Llama3.1-8B &
        Correct: $+1$ &
        $\gamma=1$ &
        \makecell{8 \\ (stopped early)} &
        8 & 512 & 1 &
        \makecell{Prompt: 2048 \\ Gen: 14336} &
        5e-7 &
        0.01 &
        0.1 \\
    \midrule
\end{tabular}
\label{fig:exp-hyperparams-reward-reinforce}
\end{table}

\subsection{Implementation of the Model-Based Verifier}\label{app:model-based-verifier}

We used \texttt{Qwen2.5-7B-Instruct} as our model-based verifier. It was provided with both the reference answer and the suffix of the long CoT. We truncated the long CoT to avoid confusing the verifier. We used the following prompt.

\begin{tcolorbox}[colback=lightgray!10, colframe=black, title={Prompt Template for Model-Based Verifier}]
\begin{lstlisting}[columns=flexible,breaklines=true,basicstyle=\small\tt]
Given the following last 20 lines of the LLM response to a math question
and the reference solution to that question, evaluate if the LLM response is correct based only on the LLM's final answer.

LLM response (last 20 lines):
...
{out}

Reference solution:
{ref}

Explain your thought process step-by-step before responding with `Judgement: <correct/wrong/not_found>`
\end{lstlisting}
\end{tcolorbox}

\subsection{Implementation of Short-Form Answer Extraction}\label{app:ans-extract}

We use the \texttt{Llama-3.1-8B-Instruct} model to extract short-form answer from QA pairs in WebInstruct, with the following prompt template:

\begin{tcolorbox}[colback=lightgray!10, colframe=black, title={Prompt Template for Short-Form Answer Extraction}]
\begin{lstlisting}[columns=flexible,breaklines=true,basicstyle=\small\tt]
Problem: {Problem}

Solution: {Solution}

Based on the Problem and the Solution, extract a short final answer that is easy to check.
Provide the short final answer in the format of "The final answer is $$
\boxed{...}
$$" 
- If the answer is a mathematical object, write it in LaTeX, e.g., "The final answer is $$
\boxed{\frac{1}{2}}
$$"
- If the answer is a boolean, write it as "True" or "False", e.g., "The final answer is $$
\boxed{True}
$$"
- If the Problem can't be answered in a short form, respond with "" like "The final answer is $$
\boxed{}
$$"
\end{lstlisting}
\end{tcolorbox}

For generation parameters, we use temperature $t=0$ (greedy decoding) and set the maximum output length as 512 tokens.

After generation, we simply extract the short-form answer from within the \texttt{\textbackslash boxed\{...\}}.

\subsection{Action Prompting Framework}
\label{app:action-prompting}

We studied the publicly released CoTs of \texttt{o1-preview} and identified that its thoughts could be categorized into a few types of actions (listed below). To construct long CoTs, we designed prompts for each of these actions and implemented a multi-step prompting framework to sequence them. The framework ceded control flow of the CoT to the LLM, with the LLM making branching or looping decisions while the framework acted more passively as a state machine reacting to the LLM outputs. The framework took care of the boilerplate around constructing the CoT with an append-only log and managed all of the orchestration.

\begin{itemize}
\item \texttt{clarify}: Making some observations about the problem in order to identify an approach to solve it.
\item \texttt{decompose}: Breaking the current problem down into smaller and easier sub-problems to solve.
\item \texttt{solution\_step}: Computing a single step in the solution. In the context of math, this could be doing some arithmetic or symbolic manipulation.
\item \texttt{reflection}: Evaluating the current approach and partial solution to see if any mistakes were made, any sub-goals were achieved, or if alternative approaches should be considered instead. Note that we used a strong teacher model \texttt{o1-mini} for the \texttt{reflection} action as that one was a more difficult prompt to respond to correctly as it requires self-correction.
\item \texttt{answer}: Responding with a final answer and terminating the CoT.
\end{itemize}

\subsubsection{Control Flow}

Simplified description of the interaction between the framework and LLM.

\begin{algorithm}[H]
\caption{Action Prompting State Machine}\label{alg:problem-solving-state-machine}
\begin{algorithmic}[1]
    \STATE {\bfseries Input:} $prompt$
    \STATE {\bfseries Output:} $chain\_of\_thought$ sequence
    \STATE $chain\_of\_thought \leftarrow \text{[$prompt$]}$ \COMMENT{Initialize singleton chain of thought sequence from prompt}
    \STATE $state \leftarrow \text{ ``clarify''}$
    \WHILE{$\text{True}$}
        \IF{$state = \text{``clarify''}$}
            \STATE $output \leftarrow \text{prompt\_action\_clarify}()$
            \STATE $(state, thought) \leftarrow \text{parse}(output)$
            \STATE $chain\_of\_thought.\text{append}(thought)$
        \ELSIF{$state = \text{``decompose''}$}
            \STATE $output \leftarrow \text{prompt\_action\_decompose}()$
            \STATE $(state, thought) \leftarrow \text{parse}(output)$
            \STATE $chain\_of\_thought.\text{append}(thought)$
        \ELSIF{$state = \text{``solution\_step''}$}
            \STATE $output \leftarrow \text{prompt\_action\_solution\_step}()$
            \STATE $(state, thought) \leftarrow \text{parse}(output)$
            \STATE $chain\_of\_thought.\text{append}(thought)$
        \ELSIF{$state = \text{``reflection''}$}
            \STATE $output \leftarrow \text{prompt\_action\_reflection}()$
            \STATE $(state, thought) \leftarrow \text{parse}(output)$
            \STATE $chain\_of\_thought.\text{append}(thought)$
        \ELSIF{$state = \text{``answer''}$}
            \STATE $output \leftarrow \text{prompt\_action\_answer}()$
            \STATE $(state, thought) \leftarrow \text{parse}(output)$
            \STATE $chain\_of\_thought.\text{append}(thought)$
            \STATE {\bfseries return} $chain\_of\_thought$ \COMMENT{Terminate after answer action}
        \ENDIF
    \ENDWHILE
\end{algorithmic}
\end{algorithm}

\subsubsection{Action Prompting Templates}

\begin{tcolorbox}[colback=lightgray!10, colframe=black, title={Action: Clarify}]
\begin{lstlisting}[columns=flexible,breaklines=true,basicstyle=\small\tt]
You are a very talented mathematics professor.
In a few sentences, VERY CONCISELY rephrase the problem to clarify its meaning and explicitly state what needs to be solved. Highlight any assumptions, constraints and potential misinterpretations.
Do NOT attempt to solve the problem yet -- you are just clarifying the problem in your mind.

<problem>
{goal}
</problem>

Answer in the following format:

<clarification>
Problem clarification as instructed above
</clarification>
<goal>
Summarize the problem into a single statement describing the goal, e.g. Find the value of the variable w.
</goal>
\end{lstlisting}
\end{tcolorbox}

\begin{tcolorbox}[colback=lightgray!10, colframe=black, title={Action: Decompose}]
\begin{lstlisting}[columns=flexible,breaklines=true,basicstyle=\small\tt]
You are a talented mathematics professor.
You already have a partial solution to a problem.
In a single sentence, propose candidates for the next subgoal as the next step of the partial solution that will help you make progress towards the current goal.
Do not repeat any subgoal, we don't want any infinite loops!
Do not suggest using a computer or software tools.

<current goal>
{current_goal}
</current goal>
<parent goal>
{parent_goal}
</parent goal>
<partial solution>
{solution}
</partial solution>

Format your answer as follows:

<thinking>
step-by-step thinking of what the next possible subgoal should be, as well as some other alternatives that might also work
remember, we want to solve the parent goal WITHOUT repeating the subgoals that are already DONE.
do not suggest verification or checking.
{parent_goal}
</thinking>
<sentence>
single sentence describing the subgoal
phrase it as if you were thinking to yourself and are considering this as a hypothesis (don't express too much certainty)
</sentence>
<sentence>
single sentence describing an *ALTERNATIVE* subgoal, without repeating previous ones
start off with "Alternatively,"
</sentence>
<sentence>
single sentence describing an *ALTERNATIVE* subgoal, without repeating previous ones
start off with "Alternatively,"
</sentence>
\end{lstlisting}
\end{tcolorbox}

\begin{tcolorbox}[colback=lightgray!10, colframe=black, title={Action: Solution Step}]
\begin{lstlisting}[columns=flexible,breaklines=true,basicstyle=\small\tt]
You are an extremely PEDANTIC mathematics professor who loves to nitpick.
You already have a partial solution to a problem. Your task is to solve *only* the current goal.
You should include symbols and numbers in every sentence if possible.

<current goal>
{current_goal}
</current goal>
<partial solution>
{solution}
</partial solution>

BE VERY CONCISE. Include calculations and equations in your response if possible, and make sure to solve them instead of just describing them.
DO NOT SOLVE THE WHOLE QUESTION, JUST THE CURRENT GOAL: {current_goal}
Do not repeat any calculations that were already in this prior step:
{prior_step}
\end{lstlisting}
\end{tcolorbox}

\begin{tcolorbox}[colback=lightgray!10, colframe=black, title={Action: Reflection}]
\begin{lstlisting}[columns=flexible,breaklines=true,basicstyle=\small\tt]
You are a talented mathematics professor.
You already have a partial solution to a math problem.
Verify whether the current subgoal has been achieved.

<current goal>
{current_goal}
</current goal>
{parent_goal}
<partial solution>
{solution}
</partial solution>

Format your answer as follows:

<verification>
Come up with a quick, simple and easy calculation to double check that the solution is correct.
This calculation should not re-compute the solution in the same way, as that would defeat the purpose of double-checking.
Use one of the following strategies:
- An easier, alternative method to arrive at the answer
- Substituting specific values into equations and checking for consistency
- Working backwards from the answer to derive the given inputs and then checking for consistency
Be consise. Do not suggest using a computer.
At the end of your verification, restate the answer from the current solution. Do not calculate it if it hasn't been solved.
Phrase it as if you are reflecting as you solve the problem.
</verification>
<current_goal_achieved>
true or false, depending on whether the solution is correct and the current goal has been achieved: {current_goal}
</current_goal_achieved>
<parent_goal_achieved>
true or false, depending on whether the parent goal has been achieved:
{parent_goal.target}
</parent_goal_achieved>
<new_goal>
If the solution is not correct or the current goal has not been achieved, suggest an alternative current goal here in a single sentence.
Start off with "Alternatively,"
Your goal should be sufficiently different from subgoals that have been solved or that have timed out:
{parent_goal_tree}
</new_goal>
\end{lstlisting}
\end{tcolorbox}

\begin{tcolorbox}[colback=lightgray!10, colframe=black, title={Action: Answer}]
\begin{lstlisting}[columns=flexible,breaklines=true,basicstyle=\small\tt]
Extract the final answer, making sure to obey the formatting instructions.
Solution:
{solution}

Formatting instructions:
{format}
\end{lstlisting}
\end{tcolorbox}
\newpage

\section{Long CoT Patterns in Pre-training Data}
\subsection{Snapshot of webpages}

Source: \href{https://brilliant.org/wiki/verify-solutions/}{brilliant.org}

The following two examples demonstrate how explicit verification after answering a question can naturally exist on a webpage.

\begin{tcolorbox}[label=webpage:explicit-revision-correct, colback=lightgray!10, colframe=black, title={Explicit verification}]
\(x+7=10\)

This problem can be solved by subtracting 7 from each side.

\(x+7-7=10-7\)

\(x=3\)

Once the problem is solved, the solution can be verified by rewriting the problem with 3 substituted for \(x\).

\(3+7=10\)

\(10=10\)

Both sides are equal, verifying that \(x=3\) is a valid solution.
\end{tcolorbox}


\begin{tcolorbox}[label=webpage:explicit-revision-wrong, colback=lightgray!10, colframe=black, title={Explicit verification that found an error}]
\(x+7=10\)
A student rushing through her homework might mistakenly write \(x=2\) as the solution to this problem. If she takes a moment to rework the equation with her answer, she will realize the answer is incorrect.

\(x+7=10\)

\(2+7=10\)

\(9=10\)

Since \(9\neq10\), the student knows she needs to go back and find a different solution to the problem.
\end{tcolorbox}

\newpage

Source: \href{https://kidswholovemath.substack.com/}{kidswholovemath.substack.com}

\begin{tcolorbox}[label=webpage:multi-solutions, colback=lightgray!10, colframe=black, title={Attempt the question from different perspective}]
\textbf{The Double Check Game} \\
Regardless of the scenario, we can play the double check game! \\
The game is simple: we try to solve the problem in as many different ways as possible.

\textbf{Elementary School Example} \\
Math problem is: $78 - 57 = ?$

To play the game, we try to solve the problem in as many different ways as possible.

\textbf{The first solution:} \\
$? = 78 - 57$ \\
Break apart the 57: \\
$? = 78 - 50 - 7$ \\
$? = 28 - 7$ \\
$? = 21$

\textbf{A second solution:} \\
$? = 78 - 57$ \\
Subtract an easier number from 78: \\
$? = 78 - 60 + 3$ \\
$? = 18 + 3$ \\
$? = 21$

\textbf{A third solution:} \\
$? = 78 - 57$ \\
Subtract 57 from an easier number: \\
$? = 80 - 57 - 2$ \\
$? = 23 - 2$ \\
$? = 21$

...
\end{tcolorbox}

\newpage

\subsection{OpenWebMath}
\label{app:open-web-math}

\subsubsection{Queries}
\label{app:open-web-math-queries}

We used \texttt{GPT-4o} to generate examples of typical pivot keywords found in long CoT. These were used to find documents in OpenWebMath that have interesting properties characteristic of long CoT trajectories.
\begin{tcolorbox}[colback=lightgray!10, colframe=black, title={"Aha" Phrases}]
\begin{verbatim}
"Let's think step by step."
"Let's go through this one step at a time."
"Breaking it down step by step..."
"Thinking about it logically, first..."
"Step 1: Let's figure out the starting point."
"If we follow the steps carefully, we get..."
"To solve this, let’s analyze it piece by piece."
"Going through this systematically, we have..."
"Okay, let’s solve this gradually."
"Does that make sense?"
"Is this correct?"
"Wait, does that check out?"
"Am I missing something?"
"Hmm… does that work?"
"Let me verify that."
"That makes sense, right?"
"Hold on, is this right?"
"Let’s double-check this."
"Wait, actually..."
"Oh, hold on..."
"Wait a second..."
"Actually, let me rethink that."
"Hmm, let me go back for a moment."
"I might need to check this again."
"Let's pause and reassess."
"Let’s check by doing the reverse."
"Let's verify by working backward."
"Can we check this by reversing the process?"
"To confirm, let's undo the steps."
"A good way to verify is by reversing it."
"If we undo the operations, do we get the same result?"
...
\end{verbatim}
\end{tcolorbox}

\newpage

\subsubsection{Matches}
\label{app:open-web-math-matches}

Source: \href{https://discourse.mc-stan.org/t/interpretation-of-multilevel-parameters/20846}{MC Stan Discussion Forum}

The discussion below took place on a message board for the probabilistic programming framework MC Stan. The user Tiny has a question about how to interpret some data and multiple other users are responding. We can see the usual pivot keywords (highlighted in \textbf{bold}) characteristic of long CoT, including branching, self-correction and even an assessment of the feasibility of an approach.

\begin{tcolorbox}[colback=lightgray!10, colframe=black, title={Discussion on message board}]
\begin{lstlisting}
So the question is then to find the right prediction task, looking at your setup, those may include:

...

• For a hypothetical future serial drawn from the same “population” as the observed serials. (i.e. include the varying intercept via a new level and “sample_new_levels = ‘uncertainty’”)
• For the “true” or “average” underlying system (i.e. ignore the varying intercept)
• In the experiments you actually observed (i.e. include the fitted varying intercepts for your experiments)

But you could also ask other stuff, like:

• What is the expected difference in some of the constants (or anything else) between two future experiments?

All of those (and more) should be answerable using the posterior of the model. But you still need to figure out which questions do you actually want to ask, as there is a lot of options…

|\textbf{Does that make sense?}|

Best of luck with your model!

...

I am not sure I follow your thought here, but |\textbf{maybe that’s just because I would have worded it differently}|?

...

|\textbf{An alternative approach would be to try}| to find a different parametrization of the model where the parameters are interpretable separately, |\textbf{but that might be hard.}|

Also, if this is the parametrization of the process used by many in the field, than maybe poeple would expect you to report as (\frac {L} {mol})^{n-1} s^{-1}, because that’s what everybody has been doing (although possibly with fixed n)?

|\textbf{Does that make sense?}|

|\textbf{Can you not just}| recast the model (with modified parameters) as

...
\end{lstlisting}
\end{tcolorbox}

\newpage

Source: \href{https://www.physicsforums.com/threads/cylinder-in-3-d.934285/}{physicsforums.com}

The discussion below took place on a physics forum. The user Songoku is asking for help with homework and another user BvU is trying to assist without revealing the solution directly. We see the usual pivot keywords indicating self-reflection, expression of uncertainty and formulation of hypotheses.

\begin{tcolorbox}[colback=lightgray!10, colframe=black, title={Discussion on a physics forum}]
\begin{lstlisting}
# Cylinder in 3 D
1. Dec 13, 2017

### songoku
1. The problem statement, all variables and given/known data
Let r be a positive constant. Consider the cylinder x2 + y2 <= r2, and let C be the part of the cylinder that satisfies
0 <= z <= y.
(1) Consider the cross section of C by the plane x = t (-r <= t <= r), and express its area in terms of r, t.
(2) Calculate the volume of C, and express it in terms of r.

...

5. Dec 13, 2017
### BvU
Simple case: x = 0. So -1 <= y <= 1. In the yz plane 0 <= z <= y is a triangle.
What about y ?

6. Dec 13, 2017
### songoku
|\textbf{I think I am missing something here}| because I feel I can't really grasp the hint given.
Let me start from the basic again:
1. Let the x - axis horizontal, y - axis vertical and z - axis in / out of page. I imagine there is circle on xy plane with radius r then it extends out of page (I take out of page as z+) to form 3 D cylinder. |\textbf{Is this correct?}|
2. Plane x = t is like the shape of a piece of paper hold vertically with the face of paper facing x - axis (I mean x - axis is the normal of the plane). |\textbf{Is this correct?}|
Thanks

7. Dec 14, 2017
### BvU
Yes

8. Dec 14, 2017
### songoku

"Consider the cross section of C by plane x = t" means plane x = t cuts the cylinder?
And the intersection will be rectangle?
...
\end{lstlisting}
\end{tcolorbox}

\newpage

Source: \href{https://math.stackexchange.com/questions/2938153/probability-that-we-stop-flipping-after-exactly-ten-flips-in-a-biased-coin-flipp/2938356}{StackExchange}

The user Baymax is asking for help on a probability problem and we see dialogue with another user Lulu. We see that the quick back-and-forth between them is similar to the kind of nimble branching behavior in long CoT where multiple solutions are quickly assessed and considered. We also see an expression of realization which can be easily re-cast as self-verification in a long CoT.  

\begin{tcolorbox}[colback=lightgray!10, colframe=black, title={Discussion on Stack Exchange}]
\begin{lstlisting}
# probability that we stop flipping after exactly ten flips in a biased coin flipping?

...

I thought that let us fix of getting a third head at last that is at 10th flip, so that we would stop there, and the remaining - getting two heads can be accommodated in the 9 trials. so there are $$9$$ choose 2 ways of getting two heads so the probability that we stop flipping after exactly ten flips is $$^9C_{2}$$ . $$\frac{1}{4}^3$$.$$\frac{3}{4}^7$$. |\textbf{Is this correct?}|

EDIT - Now the probability of getting exactly 3 heads? I got it to be $$^{10} C_{3} \frac{1}{4}^3 \frac{3}{4}^7$$. Should we get the same as the previous one? any reason why they should/should not be same?

• I think you switched $P(H),P(T)$ |\textbf{but the approach is good.}| – lulu Oct 1 '18 at 16:13
• |\textbf{oh i see now!}| thanks! – BAYMAX Oct 1 '18 at 16:13
• @lulu please see the edit – BAYMAX Oct 1 '18 at 16:30
• Your probability for exactly 3 heads is right as well. It should be obvious why the results have to be different. In the first case the outcome of the last flip is fix and in the second case the outcome of the last flip is not fix. – callculus Oct 1 '18 at 16:31

...
\end{lstlisting}
\end{tcolorbox}

\newpage

Source: \href{https://puzzling.stackexchange.com/questions/274/choosing-units-for-drug-testing}{StackExchange}

User88 interacts with multiple other users. Observe that they are helping to clarify each others' doubts, which is reminiscent of self-correction in long CoT trajectories.

\begin{tcolorbox}[colback=lightgray!10, colframe=black, title={Discussion on Stack Exchange}]
\begin{lstlisting}
Choosing units for drug testing

Here's a third puzzle that I found in a book, slightly paraphrased because I don't entirely remember the format of the original.

...

How can he arrange the dosage amounts so that he ends up using all 25 test packages, and the total units of dosage used in the tests are as low as possible?

The book had the answer, but one, it didn't explain how the answer was arrived at, and two, I don't remember what the answer was and no longer have that book with me.

• |\textbf{Am I missing something}|, or is the goal just to find 25 coprime numbers from 25 to 50? – Aza May 20 '14 at 4:33
• They don't have to be coprime. There just can't be any two where one is a factor of the other. And the range is from 1 to 50, not 25 to 50. – Joe Z. May 20 '14 at 4:34
• |\textbf{Wouldn't a single test}| of 1 unit technically satisfy the requirement? Or |\textbf{am I missing something? Ah, I guess you have to}| perform exactly 25 tests. – arshajii May 20 '14 at 14:28
• |\textbf{Yea.}| Wouldn't 1 win? – awesomepi May 20 '14 at 19:24
• You have to use all 25 tests. – Joe Z. May 20 '14 at 19:31

By logically starting from 26-50 and trying to shrink them one by one you can easily show: $8,12,14,17,18,19,20,21,22,23,25,26,27,29,30,31,33,35,37,39,41,43,45,47,49$

Which equals $711$

...
\end{lstlisting}
\end{tcolorbox}

\end{document}